\pdfoutput=1
\documentclass[11pt]{article}

\usepackage[utf8]{inputenc}
\usepackage[margin=1in]{geometry}
\usepackage{amsmath,amssymb}
\usepackage{booktabs}
\usepackage{enumitem}
\usepackage{microtype}
\usepackage{xcolor}
\usepackage[numbers,sort&compress]{natbib}
\usepackage[colorlinks=true,allcolors=blue!45!black]{hyperref}
\usepackage{authblk}
\usepackage{algorithm}
\usepackage{algpseudocode}
\usepackage{caption}
\usepackage{tikz}
\usepackage{pgfplots}
\pgfplotsset{compat=1.16}
\usetikzlibrary{arrows.meta,positioning,calc,decorations.pathreplacing,fit,backgrounds,patterns}
\usepgfplotslibrary{groupplots}
\definecolor{cOrd}{HTML}{2A78D6}
\definecolor{cGrow}{HTML}{EB6834}
\definecolor{cDorm}{HTML}{1BAF7A}
\definecolor{cRand}{HTML}{8A8983}
\definecolor{cInk}{HTML}{0B0B0B}
\definecolor{cInkTwo}{HTML}{52514E}
\definecolor{cGrid}{HTML}{E6E5E0}
\colorlet{cOrdFill}{cOrd!16}
\colorlet{cGrowFill}{cGrow!18}
\colorlet{cDormFill}{cDorm!18}
\pgfplotsset{
  dpn/.style={
    width=\linewidth, height=4.6cm,
    axis lines=left, axis line style={cInkTwo, line width=0.4pt},
    tick style={cInkTwo, line width=0.4pt}, tick align=outside, major tick length=2pt,
    tick label style={font=\scriptsize, text=cInkTwo},
    label style={font=\footnotesize, text=cInk},
    title style={font=\footnotesize\bfseries, text=cInk, yshift=-3pt},
    ymajorgrids, grid style={cGrid, line width=0.4pt},
    legend style={draw=none, fill=none, font=\scriptsize, text=cInk, legend cell align=left, inner sep=1pt, row sep=-1pt},
    every axis plot/.append style={line width=1.1pt, line join=round, line cap=round},
    clip=false,
  },
  ordline/.style={cOrd, mark=square*, mark size=1.8pt, mark options={solid, fill=cOrd, draw=white, line width=0.5pt}},
  growline/.style={cGrow, mark=*, mark size=2.0pt, mark options={solid, fill=cGrow, draw=white, line width=0.5pt}},
  dormline/.style={cDorm, mark=triangle*, mark size=2.4pt, mark options={solid, fill=cDorm, draw=white, line width=0.5pt}},
  randline/.style={cRand, dashed, mark=diamond*, mark size=2.2pt, mark options={solid, fill=cRand, draw=white, line width=0.5pt}},
  scratchpts/.style={only marks, mark=square, mark size=1.9pt, cOrd, mark options={solid, line width=0.9pt}},
}

\newcommand{\TP}{Task Path}
\newcommand{\GP}{Growth Path}
\newcommand{\act}{\bar{a}}            %
\newcommand{\lev}{\bar{n}}            %
\newcommand{\thdie}{\theta_{\mathrm{die}}}
\newcommand{\thdiv}{\theta_{\mathrm{div}}}
\newcommand{\live}{\mathcal{A}}       %

\newcommand{\papertitle}{The Growth Path Is the Missing Half of Neural Networks}

\title{\papertitle}
\newcommand{\samethanks}[1][\value{footnote}]{\footnotemark[#1]}
\author[1]{Tingan Jin\thanks{Equal contribution.}}
\author[2]{Haosong Li\samethanks}
\affil[1]{University of California, Los Angeles}
\affil[2]{Independent Researcher}
\date{}

\begin{document}
\maketitle

\begin{abstract}
An artificial neuron computes and learns, but it does not live: the units of a network are fixed when it is designed, and every one of them is kept and paid for whether it is used or not.
We give each neuron a second pathway.
Beside its \emph{\TP{}}, the usual computation and gradient learning, a \emph{dual-pathway neuron} has a \emph{\GP{}}: a nutrient flows from the outputs back through the network, every layer receives the same fixed budget, and each unit is fed in proportion to how much the units above it use it.
A well-fed unit divides and a starving unit dies.
This allocation is not a separate analysis or pruning step: it runs for as long as the network trains, as a standing rule that reads only activations and weights.
The task is still learned from labels with the ordinary loss, and the training procedure gains no phase, no schedule and no target size.
Because the budget is conserved, the competition is zero-sum and a single number sets the size of the network.
Switched on in a pretrained network that is deployed on a narrower domain, the \GP{} shrinks the network while it keeps learning.
With Adam the network ends 1.0 to 1.2 points more accurate on the domain than the same network trained on in the ordinary way, at 24\% to 60\% of its compute, and at 12\% of the compute it is still ahead; a random choice of the same number of units gives no gain.
Letting starving pretrained units sleep instead of die keeps more of the rarely used content in every setting, by 1.2 to 4.6 points.
On long streams of changing tasks a network with the \GP{} holds its accuracy over 800 tasks while ordinary networks decline.
With plain SGD the gains are smaller: at deployment the method trades compute against rarely used content, and on one of the two continual-learning streams it only ties networks of equal size. We report both optimisers side by side.
The experiments are small (three-layer perceptrons on MNIST-like data); what they show is that a conserved, loss-free second pathway can decide by itself which units a network needs.
\end{abstract}

\section{Introduction}\label{sec:intro}

An artificial neuron has one job.
It computes an activation, and it adjusts its weights to reduce a loss.
Whether it should exist is not its concern: the number of units is chosen by a designer before training starts, and every unit is then kept for the life of the network, used or not.
A unit that has stopped contributing still occupies memory and still costs a multiplication at every step.
In one of the networks we study below, more than nine tenths of the units in two of the three hidden layers are inactive, and all of them are computed at every step.

A biological neuron has a second job: to stay alive.
During development far more neurons are produced than are kept, and they compete for trophic factors that their target cells supply in limited amounts; neurons that secure too little die \citep{levimontalcini1987nerve,oppenheim1991cell,raff1992social}.
Learning and survival run in the same cell, on different signals.
Networks of artificial neurons have the first half of this arrangement and lack the second.
Methods that change the structure of a network do exist (growing, pruning, resetting of units), but they act on the network from outside: a separate procedure, driven by the loss, by gradients or by a utility score, with a schedule, a quota or a target size chosen by the designer (Section~\ref{sec:related}).

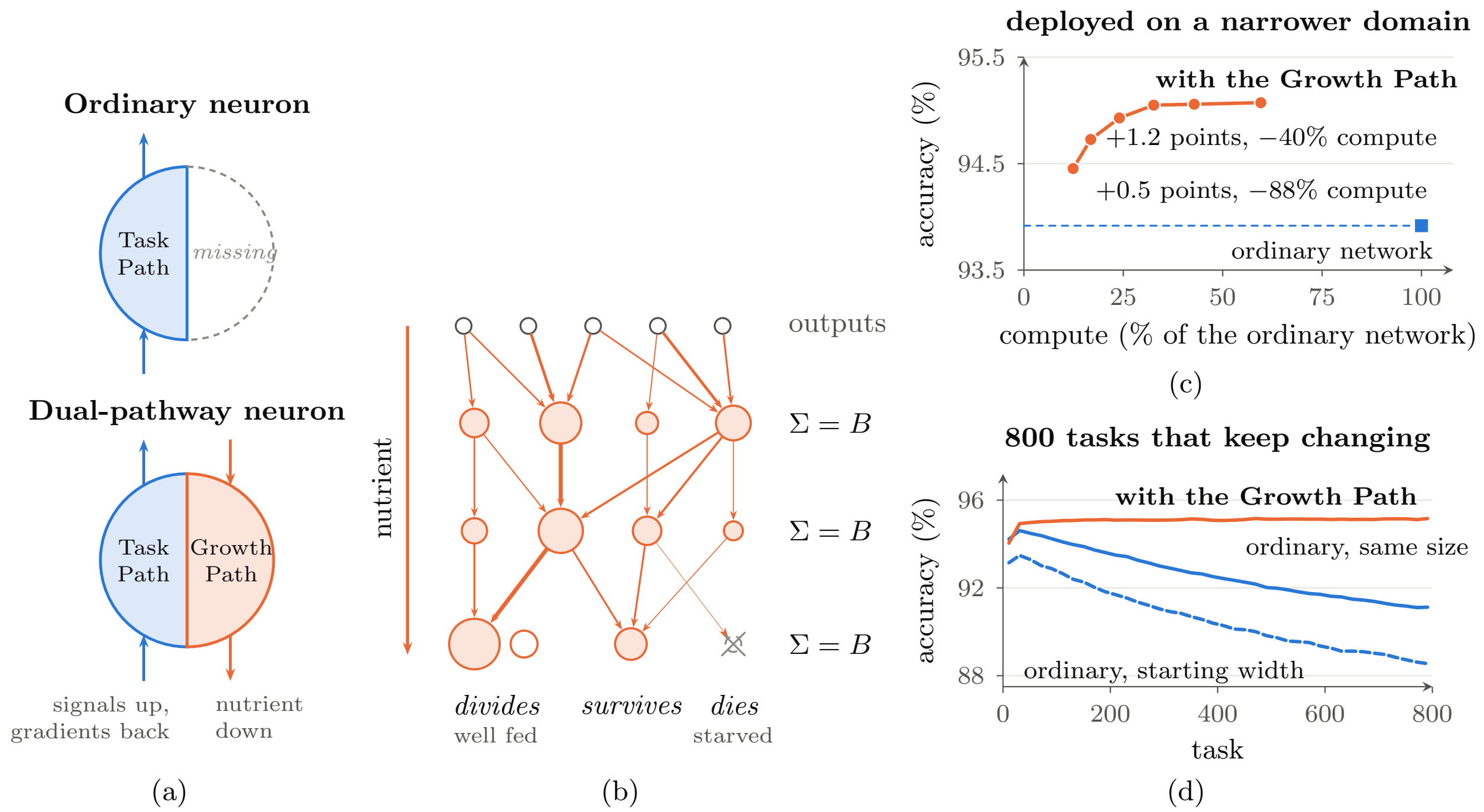
\begin{figure}[t]
  \centering
  \begin{minipage}[b]{0.23\linewidth}\centering
\begin{tikzpicture}[x=1.12cm,y=1.12cm,font=\footnotesize]
  \def\r{0.86}
  \tikzset{half/.style={font=\fontsize{7.5}{8.5}\selectfont, align=center, text=cInk},
           tp/.style={cOrd, -{Stealth[length=4pt]}, line width=0.9pt},
           gp/.style={cGrow, -{Stealth[length=4pt]}, line width=0.9pt}}
  \begin{scope}[shift={(0,3.05)}]
    \fill[cOrdFill] (90:\r) arc (90:270:\r) -- cycle;
    \draw[cOrd, line width=0.9pt] (90:\r) arc (90:270:\r) -- cycle;
    \draw[cRand, line width=0.7pt, dash pattern=on 2pt off 1.6pt] (90:\r) arc (90:-90:\r);
    \node[half] at (-0.43,0) {Task\\Path};
    \node[half, text=cRand, font=\fontsize{7.5}{8.5}\selectfont\itshape] at (0.47,0) {missing};
    \draw[tp] (-0.43,-1.2) -- (-0.43,-\r+0.12);
    \draw[tp] (-0.43,\r-0.12) -- (-0.43,1.2);
    \node[text=cInk, font=\footnotesize\bfseries, anchor=south] at (0,1.22) {Ordinary neuron};
  \end{scope}
  \begin{scope}[shift={(0,0)}]
    \fill[cOrdFill] (90:\r) arc (90:270:\r) -- cycle;
    \fill[cGrowFill] (90:\r) arc (90:-90:\r) -- cycle;
    \draw[cOrd, line width=0.9pt] (90:\r) arc (90:270:\r) -- cycle;
    \draw[cGrow, line width=0.9pt] (90:\r) arc (90:-90:\r) -- cycle;
    \node[half] at (-0.43,0) {Task\\Path};
    \node[half] at (0.44,0) {Growth\\Path};
    \draw[tp] (-0.43,-1.2) -- (-0.43,-\r+0.12);
    \draw[tp] (-0.43,\r-0.12) -- (-0.43,1.2);
    \draw[gp] (0.43,1.2) -- (0.43,\r-0.12);
    \draw[gp] (0.43,-\r+0.12) -- (0.43,-1.2);
    \node[text=cInk, font=\footnotesize\bfseries, anchor=south] at (0,1.22) {Dual-pathway neuron};
    \node[text=cInkTwo, anchor=north east, align=right, font=\fontsize{7.5}{8.5}\selectfont] at (-0.05,-1.23) {signals up,\\gradients back};
    \node[text=cInkTwo, anchor=north west, align=left, font=\fontsize{7.5}{8.5}\selectfont] at (0.16,-1.23) {nutrient\\down};
  \end{scope}
\end{tikzpicture}\\[1pt]
{\footnotesize (a)}
\end{minipage}\hfill
\begin{minipage}[b]{0.35\linewidth}\centering
\begin{tikzpicture}[x=1.2cm,y=1.2cm,font=\footnotesize]
  \tikzset{unit/.style={circle, draw=cGrow, fill=cGrowFill, line width=0.7pt, inner sep=0pt},
           outu/.style={circle, draw=cInkTwo, fill=white, line width=0.6pt, inner sep=0pt, minimum size=5pt},
           flow/.style={cGrow, -{Stealth[length=3pt]}, line cap=round}}
  \foreach \i in {0,...,4} { \node[outu] (o\i) at (0.40+\i*0.60,4.05) {}; }
  \node[unit, minimum size=9pt]  (c0) at (0.50,3.15) {};
  \node[unit, minimum size=13pt] (c1) at (1.30,3.15) {};
  \node[unit, minimum size=7pt]  (c2) at (2.10,3.15) {};
  \node[unit, minimum size=11pt] (c3) at (2.90,3.15) {};
  \node[unit, minimum size=8pt]  (b0) at (0.50,2.15) {};
  \node[unit, minimum size=14pt] (b1) at (1.30,2.15) {};
  \node[unit, minimum size=9pt]  (b2) at (2.10,2.15) {};
  \node[unit, minimum size=6pt]  (b3) at (2.90,2.15) {};
  \node[unit, minimum size=16pt] (a0) at (0.50,1.10) {};
  \node[unit, minimum size=8.5pt, fill=white] (a0c) at (0.96,1.10) {};
  \node[unit, minimum size=10pt] (a2) at (1.95,1.10) {};
  \node[circle, draw=cRand, fill=white, line width=0.6pt, densely dashed, inner sep=0pt, minimum size=5pt] (a3) at (2.90,1.10) {};
  \draw[cRand, line width=0.6pt] ($(a3)+(-0.11,-0.11)$) -- ($(a3)+(0.11,0.11)$) ($(a3)+(-0.11,0.11)$) -- ($(a3)+(0.11,-0.11)$);
  \foreach \o/\c/\w in {o0/c0/0.5,o0/c1/0.5,o1/c1/0.9,o2/c1/0.7,o2/c3/0.5,o3/c3/0.9,o4/c3/0.6,o3/c2/0.35}
    { \draw[flow, line width=\w pt] (\o) -- (\c); }
  \foreach \c/\b/\w in {c0/b0/0.5,c0/b1/0.4,c1/b1/1.3,c3/b1/0.7,c3/b2/0.7,c2/b2/0.35,c3/b3/0.3}
    { \draw[flow, line width=\w pt] (\c) -- (\b); }
  \foreach \b/\a/\w in {b0/a0/0.6,b1/a0/1.3,b1/a2/0.6,b2/a2/0.6,b3/a2/0.3,b2/a3/0.15}
    { \draw[flow, line width=\w pt] (\b) -- (\a); }
  \foreach \y in {3.15,2.15,1.10} { \node[text=cInk, anchor=west] at (3.30,\y) {$\Sigma = B$}; }
  \node[text=cInkTwo, anchor=west] at (3.30,4.05) {outputs};
  \node[text=cInk, anchor=north, font=\footnotesize\itshape] at (0.70,0.74) {divides};
  \node[text=cInk, anchor=north, font=\footnotesize\itshape] at (1.95,0.74) {survives};
  \node[text=cInk, anchor=north, font=\footnotesize\itshape] at (2.90,0.74) {dies};
  \node[text=cInkTwo, anchor=north, font=\fontsize{7.5}{8.5}\selectfont] at (0.70,0.44) {well fed};
  \node[text=cInkTwo, anchor=north, font=\fontsize{7.5}{8.5}\selectfont] at (2.90,0.44) {starved};
  \draw[cGrow, -{Stealth[length=4pt]}, line width=1pt] (-0.12,4.05) -- (-0.12,1.0);
  \node[text=cInk, rotate=90, anchor=south] at (-0.16,2.55) {nutrient};
\end{tikzpicture}\\[1pt]
{\footnotesize (b)}
\end{minipage}\hfill
\begin{minipage}[b]{0.385\linewidth}\centering
\begin{tikzpicture}
\begin{axis}[dpn, height=3.95cm, width=\linewidth,
  title={deployed on a narrower domain},
  xmin=0, xmax=108, ymin=93.5, ymax=95.5,
  xtick={0,25,50,75,100}, ytick={93.5,94.5,95.5},
  xlabel={compute (\% of the ordinary network)}, ylabel={accuracy (\%)},
  xlabel style={yshift=2pt}, ylabel style={yshift=-2pt}]
  \addplot[cOrd, line width=0.6pt, dash pattern=on 2pt off 2pt, mark=none, forget plot] coordinates {(0,93.917) (100,93.917)};
  \addplot[growline] table[x=rel,y=common] {fig/data/cost_adam_death.dat};
  \addplot[ordline, only marks, mark size=2.3pt] table[x=rel,y=common] {fig/data/ordinary_adam.dat};
  \node[font=\scriptsize, text=cInk, anchor=north east] at (axis cs:106,93.89) {ordinary network};
  \node[font=\scriptsize\bfseries, text=cInk, anchor=south west] at (axis cs:30,95.10) {with the Growth Path};
  \node[font=\scriptsize, text=cInk, anchor=north east] at (axis cs:107,94.97) {$+1.2$ points, $-40\%$ compute};
  \node[font=\scriptsize, text=cInk, anchor=north west] at (axis cs:15,94.47) {$+0.5$ points, $-88\%$ compute};
\end{axis}
\end{tikzpicture}\\[-4pt]
{\footnotesize (c)}\\[4pt]
\begin{tikzpicture}
\begin{axis}[dpn, height=3.95cm, width=\linewidth,
  title={800 tasks that keep changing},
  xmin=0, xmax=800, ymin=87.5, ymax=97.2,
  xtick={0,200,400,600,800}, ytick={88,92,96},
  xlabel={task}, ylabel={accuracy (\%)},
  xlabel style={yshift=2pt}, ylabel style={yshift=-2pt}]
  \addplot[cOrd, mark=none, dash pattern=on 3pt off 1.5pt] table[x=task,y=start] {fig/data/long_adam_n300.dat};
  \addplot[cOrd, mark=none] table[x=task,y=peak] {fig/data/long_adam_n300.dat};
  \addplot[cGrow, mark=none] table[x=task,y=growth] {fig/data/long_adam_n300.dat};
  \node[font=\scriptsize\bfseries, text=cInk, anchor=south east] at (axis cs:795,95.25) {with the Growth Path};
  \node[font=\scriptsize, text=cInk, anchor=south west] at (axis cs:430,92.75) {ordinary, same size};
  \node[font=\scriptsize, text=cInk, anchor=north east] at (axis cs:585,89.25) {ordinary, starting width};
\end{axis}
\end{tikzpicture}\\[-4pt]
{\footnotesize (d)}
\end{minipage}
  \caption{\textbf{The \GP{} is a second pathway in every neuron.}
  (a)~An ordinary unit has a \TP{} only: signals go up and gradients come back. A dual-pathway neuron also receives a nutrient from the units above it.
  (b)~The nutrient enters at the outputs and is passed down; every layer receives the same budget~$B$, split among its units in proportion to their use by the layer above. A well-fed unit divides, a starved unit dies, the others survive. The allocation is repeated every 100 updates while the network trains, and it never reads the loss.
  (c)~Switched on in a pretrained network that is deployed on a narrower domain, the \GP{} shrinks the network while it keeps learning. Each point is one budget ($B=25$ to 100). At 12\% of the compute the network is still more accurate on the domain than the ordinary network at full size.
  (d)~Trained from scratch on 800 tasks that keep rewiring the input, a network with the \GP{} holds its accuracy, while ordinary networks of the same size and of the starting width decline.
  Panels (c) and (d) are for Adam (means of five seeds); the results with SGD, which are less favourable at deployment, are in Sections~\ref{sec:deployment} and~\ref{sec:continual}.}
  \label{fig:teaser}
\end{figure}

\paragraph{The missing half.}
We propose to put the second job into the neuron itself.
A \emph{dual-pathway neuron} (Figure~\ref{fig:teaser}a) has two pathways.
Its \emph{\TP{}} is everything an ordinary unit does.
Its \emph{\GP{}} is new.
A \emph{nutrient} enters the network at the outputs and flows backwards through the layers.
Every layer receives the same amount, the \emph{budget}~$B$, and each unit passes on what it receives to the units below it, in proportion to how much it uses them (Figure~\ref{fig:teaser}b).
A unit that stays well fed \emph{divides}: it hands part of its weights and of its nutrient to a child.
A unit that starves \emph{dies}.
The others \emph{survive}.
This allocation is not an analysis made once on a finished network.
It is repeated every 100 updates for as long as the network trains, and the units live on it.
Three properties set this apart from growing and pruning as they are usually done.
\begin{itemize}[nosep,leftmargin=1.4em]
  \item \textbf{It runs inside ordinary training.} The rule reads activations and weights; it sees no label, no loss value and no gradient. The task is still learned from labels with the usual cross-entropy loss, so the weights are decided by one signal and the structure by another. Nothing is added that needs attention while the network trains: no pruning phase, no retraining and no schedule (Algorithm~\ref{alg:training}).
  \item \textbf{It is conserved.} Every layer receives the same budget at every check, so the competition inside a layer is zero-sum: a unit gains only what other units lose. There is no replacement quota and no target size. Without conservation the same rule does not control the size (Section~\ref{sec:continual}).
  \item \textbf{One number sets the size.} The budget is the single knob for the size of the network, and through it for the compute. It is set once.
\end{itemize}
The way the nutrient is passed down is borrowed: allocations of this form explain the predictions of a trained network~\citep{bach2015pixel} and have served to prune one~\citep{yeom2021pruning}, and units have been grown, pruned and replaced in many ways before.
What is new, to our knowledge, is to turn this conserved allocation across layers into a growth rule that runs throughout training and decides division and death alike (Section~\ref{sec:related}).

\paragraph{What it buys.}
The clearest test is deployment (Section~\ref{sec:deployment}).
A network is pretrained on five kinds of content and then deployed on a domain that keeps two of them, shows a third only rarely, and drops the rest.
We switch the \GP{} on in the pretrained network with a budget below its width, and change nothing else.
The network shrinks within about 200 tasks and goes on learning (Figure~\ref{fig:teaser}c).
With Adam it ends 1.0 to 1.2 points more accurate on the domain than the same network trained on in the ordinary way, at 24\% to 60\% of the compute; at 12\% of the compute it is still 0.5 points ahead.
The choice of units carries the result: with the same number of deaths and divisions applied to randomly chosen units, the gain is gone.
And the units that die are disproportionately the ones that served content the domain no longer shows, although the rule never sees a label.
When starving pretrained units are put to sleep instead of removed, more of the rarely used content is kept in every setting (by 3.8 to 4.6 points with Adam and by 1.2 to 2.2 points with SGD), and dormant units wake ten to twenty times more often on the tasks that need them.

The same rule helps a network to keep learning (Figure~\ref{fig:teaser}d; Section~\ref{sec:continual}).
On a stream of 800 tasks in which old knowledge expires quickly, a network with the \GP{} holds 94\% to 95\% from the first task to the last, while ordinary networks of the same size fall from 94\% to 91\%.
On a stream in which old knowledge stays useful, it beats ordinary networks of the same size by 0.8 to 1.6 points with Adam.

\paragraph{What it does not show.}
With plain SGD the ordinary networks on our streams degrade little, and the picture changes.
At deployment the \GP{} then matches the ordinary network on the common content at a third of the compute, but it loses rarely used content, and on the second continual-learning stream it only ties ordinary networks of the same size.
All experiments use three-layer perceptrons on MNIST-like data, with five seeds per comparison.
We report both optimisers side by side in every table and collect the limits in Section~\ref{sec:limits}.
The title states a position, and the experiments are a first test of it: they show that a conserved second pathway, blind to the loss, can decide which units a network needs, and that where ordinary training wastes units it does so to the network's advantage.

\paragraph{Contributions.}
(i)~The dual-pathway neuron and its \GP{}: a conserved allocation across layers, turned into a growth rule that runs throughout training and decides division, survival and death unit by unit, while the task is learned exactly as before (Section~\ref{sec:method}).
(ii)~A deployment result: a pretrained network that shrinks to a budget while it learns a domain, with full cost--accuracy curves for two optimisers, and dormancy as a way to keep rarely used knowledge (Section~\ref{sec:deployment}).
(iii)~Evidence on the mechanism: random-turnover, no-conservation and same-size controls, and a direct look at which units die (Sections~\ref{sec:deployment}--\ref{sec:analysis}).
(iv)~A protocol in which baselines are tuned per size, rules are fixed before the check seeds are run, and unfavourable results are reported next to the favourable ones (Section~\ref{sec:protocol}).

\section{Dual-Pathway Neurons}\label{sec:method}

A dual-pathway neuron is an ordinary hidden unit with a second pathway added to it.
The \emph{\TP} is what an ordinary unit already has: its incoming weights and bias, its activation, and the gradient updates driven by the task loss.
The \emph{\GP} is a small state kept by each unit, together with a rule that decides from this state whether the unit \emph{divides}, \emph{survives} or \emph{dies}.
It reads activations and weights only, never the loss, the labels or a gradient.
The two pathways meet in one place: when a unit divides or dies, the set of weights that the \TP{} trains changes.
Algorithm~\ref{alg:training} shows where the \GP{} enters the training loop.
The loss, the labels, the data and the order of training are those of an ordinary network; the \GP{} adds a running average per unit and, every $K$ updates, one pass from the outputs down.
There is no separate phase for growing or pruning, and no step at which training has to be stopped, inspected or restarted.

\begin{algorithm}[t]
\caption{Training a network of dual-pathway neurons. The \TP{} steps are those of ordinary training. The \GP{} adds the steps marked \textcolor{cGrow}{$\blacktriangleright$}; they read activations and weights, never $y$ or the loss.}
\label{alg:training}
\begin{algorithmic}[1]
\For{each minibatch $(x,y)$ of the stream}
  \State forward pass; loss $=$ cross-entropy between the outputs and $y$ \Comment{\TP{}}
  \State backward pass; optimiser step on the weights between live units \Comment{\TP{}}
  \State \textcolor{cGrow}{$\blacktriangleright$} update the activity $\act_i$ of every live unit from its activations \Comment{\GP{}}
  \If{the number of updates is a multiple of $K$} \Comment{structural check}
    \State \textcolor{cGrow}{$\blacktriangleright$} route the budget $B$ from the outputs down, \eqref{eq:intake}; update the levels $\lev_i$
    \State \textcolor{cGrow}{$\blacktriangleright$} established units with $\lev_i<\thdie$ die; the best-fed units with $\lev_i>\thdiv$ divide, \eqref{eq:child}--\eqref{eq:pay}
  \EndIf
\EndFor
\end{algorithmic}
\end{algorithm}

We describe the method for the networks used in this paper: multilayer perceptrons with $L=3$ hidden layers of ReLU units and $M=10$ outputs.
Hidden layer $\ell$ holds a set $\live_\ell$ of $N_\ell$ \emph{live} units; $a_i(x)\ge 0$ is the activation of hidden unit $i$, and $W_{ij}$ is the weight from unit $i$ to unit $j$ of the next layer.

\subsection{Task Path}\label{sec:taskpath}

The \TP{} is ordinary learning: the loss is the cross-entropy, and the weights between live units are updated by minibatch SGD or Adam~\citep{kingma2015adam}.
A unit that is not live takes no part in the forward or the backward pass and receives no update.
Between two structural events, training differs from that of an ordinary network made of the same live units in one respect only: units that were born recently use a larger step size (Section~\ref{sec:growthpath}).

\subsection{Growth Path: a conserved nutrient routed from the outputs}\label{sec:growthpath}

\paragraph{State and structural check.}
Each hidden unit $i$ keeps three numbers: its \emph{activity} $\act_i$, an exponential moving average of its batch-mean activation (time constant 1,000 updates); its \emph{nutrient level} $\lev_i$, defined below; and its \emph{age}, the number of updates since its birth.
Every $K=100$ updates the \GP{} runs a \emph{structural check}: it routes the nutrient, updates the levels, and applies deaths and divisions.

\paragraph{Routing and conservation.}
Every layer receives the same amount of nutrient at each check, the \emph{budget} $B$.
Each output unit emits a \emph{supply} of $B/M$.
A unit $j$ passes its whole supply $\sigma_j$ to the live units of the layer below, in proportion to its \emph{use} of them, $u_{ij}=\act_i\,|W_{ij}|$, the average absolute contribution of $i$ to the input of $j$.
The \emph{intake} of a unit $i$ of layer $\ell$ is
\begin{equation}
  n_i = \sum_{j} \sigma_j\,\frac{u_{ij}}{\sum_{k\in\live_\ell} u_{kj}} ,
  \label{eq:intake}
\end{equation}
where $j$ runs over the live units of layer $\ell+1$, or over the outputs if $\ell=L$.
The supply of a hidden unit is its own intake at this check, $\sigma_i=n_i$, so the routing is a single pass from the outputs down to the first hidden layer.
The shares handed out by a downstream unit sum to one, so the intakes of every hidden layer sum to the budget:%
\footnote{This holds whenever every downstream unit with a positive supply uses at least one live unit of the layer below. If it does not, its supply is not routed at this check, and the implementation counts the event.}
\begin{equation}
  \sum_{i\in\live_\ell} n_i = B \qquad \text{for } \ell=1,\dots,L .
  \label{eq:conservation}
\end{equation}
The redistribution has the form of the conservative rules of layer-wise relevance propagation~\citep{bach2015pixel} and excitation backprop~\citep{zhang2016top}; what differs is its start (an equal supply at every output, not the score of a class) and its use (Section~\ref{sec:related}).
Two properties follow.
The competition inside a layer is zero-sum: a unit can gain intake only if other units of its layer lose it.
And the nutrient is top-down: a unit is fed for being used by units that are themselves fed, so a unit whose consumers are starving receives little, however active it is.

\paragraph{Nutrient level.}
The level is a smoothed history of the intake: at each check $\lev_i \leftarrow \lev_i + \beta\,(n_i-\lev_i)$ with $\beta = 1-e^{-K/\tau}$, a time constant of $\tau$ updates.
What is conserved is the nutrient routed at each check; the levels are a running memory of it.

\paragraph{Divide, survive or die.}
A unit is \emph{established} once its age reaches the \emph{protection period} $P$.
At each check an established unit \emph{dies} if $\lev_i<\thdie$, may \emph{divide} if $\lev_i>\thdiv$, and \emph{survives} otherwise; a unit inside its protection period survives.
Deaths are applied first.
Then at most $\max(1,\lceil\rho N_\ell\rceil)$ units of each layer divide, those with the highest levels first.
There is no limit on deaths, no minimum size and no replacement quota: the size of a layer is whatever the competition leaves.

\paragraph{Death.}
A dying unit is removed: its weights, bias and optimiser state are set to zero and its slot is freed.
The level it still holds is not handed on; it disappears, so the total level of a layer dips after deaths and recovers over the following checks.

\paragraph{Division.}
A dividing unit $p$, the parent, stays, and a child $c$ is created in a free slot of the same layer.
With $w_p$ the incoming weights and bias of the parent, the child receives
\begin{equation}
  w_c = \lambda\,\bigl(w_p + \varepsilon\,s_p\,\xi\bigr) + (1-\lambda)\,w_0 ,
  \label{eq:child}
\end{equation}
where $\lambda\in[0,1]$ is the \emph{inheritance ratio}, $\xi$ is standard Gaussian noise, $s_p$ is the root-mean-square entry of $w_p$, $\varepsilon$ is a relative noise scale, and $w_0$ is a fresh draw from the distribution used to initialise the network.
The parent pays for the child out of its outgoing weights and its level:
\begin{equation}
  W_{cj} = \tfrac{\lambda}{2}\,W_{pj},\quad
  W_{pj} \leftarrow \bigl(1-\tfrac{\lambda}{2}\bigr)W_{pj},\qquad
  \lev_c = \tfrac{\lambda}{2}\,\lev_p,\quad
  \lev_p \leftarrow \bigl(1-\tfrac{\lambda}{2}\bigr)\lev_p .
  \label{eq:pay}
\end{equation}
A division thus moves level from the parent to the child and leaves the total level of the layer unchanged.
At $\lambda=1$ the child is a noisy copy that shares the parent's outgoing weights, as in the widening step of Net2Net~\citep{chen2015net2net}.
At $\lambda=0$ it is a freshly initialised unit with zero outgoing weights and zero level, which has to earn its intake while it is protected.
In both cases the function of the network is unchanged at the division (at $\lambda=1$ up to the noise); in between it changes slightly.

\paragraph{Protection, capacity and settings.}
A newborn unit can neither die nor divide for $P$ updates, and during this period the step size of its incoming weights and bias is multiplied by $\kappa\ge 1$, so that it can find a use before it is judged.
Storage holds at most $C$ units per layer; a run in which a layer reaches $C$, loses all its units or produces non-finite values is invalid and is not used.
All deployment experiments use $\thdie=0.05$, $\thdiv=1.5$, $\tau=P=3{,}750$ updates, $\kappa=3$, $\varepsilon=0.1$, $\rho=0.01$, $\lambda=0.5$ and $C=768$; these constants were selected on tuning seeds of another stream (Section~\ref{sec:seeds}).
Appendix~\ref{app:method} gives the remaining details.

\subsection{The budget sets the size}\label{sec:budget}

The thresholds are fixed numbers, and by \eqref{eq:conservation} the mean intake in a layer is $B/N_\ell$.
In a crowded layer the shares are small and more units fall below $\thdie$; in a sparse layer more units rise above $\thdiv$.
For fixed thresholds, the budget is therefore the one quantity that sets the scale of a layer.
When a network is trained from scratch, we set $B$ to its starting width $N_0$, so the mean intake starts at 1 and the thresholds read as multiples of an equal share.
When the \GP{} is switched on in a pretrained network of width $N_0$, we choose $B<N_0$: every pretrained unit starts as an established unit with level $B/N_0$, and the network has to fit into a budget that is smaller than its width.
Whether it does, and what this costs in accuracy, is the question of Section~\ref{sec:deployment}.

Conservation also bounds the size.
The total level of a layer never exceeds $B$: a smoothing step moves it towards $B$, a division leaves it unchanged, and a death lowers it.
After a check every established unit holds at least $\thdie$, so a layer has at most $B/\thdie$ established units, whatever the length of the stream.
The bound is loose and does not predict the size, but without conservation there is none (control in Section~\ref{sec:controls}).

\subsection{Two variants: two parents, and dormancy}\label{sec:twoparents}\label{sec:dormancy}

\paragraph{Division with two parents.}
A child made from one parent at $\lambda>0$ starts as a near copy.
In the two-parent variant a second parent is drawn among the other established units of the layer with probability proportional to their levels, and the child's incoming weights combine the two, either by copying each weight from one of the parents at random or by averaging.
Each parent gives a quarter of its outgoing weights and of its level where a single parent gives a half, so the total level of the layer is again unchanged.

\paragraph{Dormancy instead of death.}
Death is irreversible: a unit that is needed only rarely can starve between two uses, and what it has learned is then lost.
For a pretrained network we therefore also study a variant in which a starving pretrained unit becomes \emph{dormant} where it would otherwise die.
It takes no part in the forward or the backward pass; its weights, optimiser state and \GP{} state are stored unchanged.
At every check each dormant unit $d$ is asked what it would receive if it alone were added back: with $\hat a_d$ its mean activation on the last training batch,
\begin{equation}
  \hat n_d = \sum_{j} \sigma_j\,\frac{\hat a_d\,|W_{dj}|}{\sum_{k\in\live_\ell} u_{kj} + \hat a_d\,|W_{dj}|} .
  \label{eq:wake}
\end{equation}
The unit wakes if $\hat n_d \ge B/N_\ell$, the mean intake of the live units of its layer, and then cannot go dormant again for one protection period.
The test adds no parameter.
Units born after the switch-on die as before; extending dormancy to them needs a rule for deleting units that never return, which we leave open.

\subsection{Counting cost}\label{sec:cost}

We count cost in parameters that take part in the computation.
The \emph{compute cost} of a network is the number of weights and biases between its live units; with dormant units it includes the wake test, counted once per check.
The \emph{storage cost} counts live and dormant units together.
We quote costs relative to an ordinary network: at deployment as a percentage of the ordinary 784--300--300--300--10 network, and on the streams trained from scratch in units of the 784--100--100--100--10 network (99,710 parameters).
The count leaves out the bookkeeping of the \GP{} itself and is not a wall-clock measurement (Section~\ref{sec:analysis}).

\section{Experimental protocol}\label{sec:protocol}

The comparisons, metrics and decision rules below were fixed before the method was run on the seeds used for conclusions.
Appendix~\ref{app:protocol} gives the full details.

\paragraph{Baseline: an ordinary network.}\label{sec:baseline}
The baseline throughout is an \emph{ordinary network}: the same perceptron with a fixed structure, trained by backpropagation on the same stream with the same optimiser, with a step size tuned separately for every size at which it is used.
A result counts in favour of the method only relative to this network.
When a network is trained from scratch the method chooses its own size, so the ordinary network is run at two matched sizes, the \emph{final size} of the method and its \emph{peak size}, and the method has to beat both.
At deployment the baseline is the same pretrained network, trained on at its fixed size.

\paragraph{Three controls.}\label{sec:controls}
\emph{Random turnover} replays a run with the same number of deaths and divisions at every check and in every layer, but draws the units at random among the eligible ones; it separates \emph{which} units are replaced from how many are replaced and when.
\emph{No conservation} feeds each unit in proportion to its own use, without the normalisation in \eqref{eq:intake}, so that the total of a layer is free to rise or fall.
A \emph{same-size ordinary network trained from scratch} asks whether size alone explains a result; at deployment it is a reference only, because it needs the target size in advance.

\paragraph{Seeds and rules fixed in advance.}\label{sec:seeds}
Every choice (structural settings, step sizes) is made on tuning seeds 0--4 by a fixed rule and then frozen.
Each comparison is run once on five \emph{check seeds} that no tuning run has used.
Stream lengths, primary metrics and verdict thresholds were written down before the check seeds were run; readings that were not fixed in advance are marked where they appear.
At deployment nothing of the method was tuned: its settings are those selected on the mixed stream, and its step size is that of the pretrained network.
We have not run a further confirmation on fresh seeds; every result rests on the five check seeds of its comparison.
We use SGD without momentum and Adam~\citep{kingma2015adam} at its default settings, with batch size 16, and compare only runs with the same optimiser.
Differences are paired by seed and reported as means with 95\% intervals (Student's $t$, 4 degrees of freedom); we call a difference positive or negative only if its interval excludes zero.

\paragraph{Three task streams.}\label{sec:streams}
A stream is a sequence of ten-class tasks that follow one another without any signal at the boundary; all networks have 784 inputs and three hidden layers.
In the \emph{pixel-permutation stream} each task is MNIST~\citep{lecun1998gradient} under a new random permutation of the pixels~\citep{goodfellow2013empirical} (200 tasks, and 800 in a long version).
The score is the online accuracy, and the primary metric is its mean over the last tenth of the tasks.
Every task rewires the input, so this stream tests whether a network can keep learning and gives no credit for keeping what was learned.
The \emph{mixed stream} has 2,000 tasks of 24,000 images each, whose ten classes are assigned to the outputs anew in every task.
Its \emph{home} content returns throughout: handwritten digits, letters~\citep{cohen2017emnist} and Kuzushiji characters~\citep{clanuwat2018kmnist}.
Two unrelated kinds visit once, for 200 tasks each: clothing~\citep{xiao2017fashion} and a \emph{second sensor}, the home images under one fixed pixel permutation.
The score of a task is its \emph{block score}, the mean held-out accuracy at four points of the task, and the primary metric is the mean block score of the last tenth of the home tasks.
The \emph{deployment stream} (Figure~\ref{fig:deploy-time}, top) has tasks of the same format.
In its \emph{general phase} (tasks 1--500) the five kinds appear with equal probability.
In its \emph{deployment phase} (tasks 501--2,000) digits and letters are the \emph{common} content, Kuzushiji is the \emph{rare} content, exactly one task in every block of 20, and clothing and the second sensor are \emph{gone}.
A network with three hidden layers of 300 units is trained through the general phase as an ordinary network; at the end of task 500 the \GP{} is switched on in it with a budget between 25 and 100.

\paragraph{Metrics and verdict rules.}\label{sec:verdicts}
On the streams trained from scratch the method passes if its primary metric is above both size-matched ordinary networks, with both intervals above zero.
For deployment, the \emph{common-content accuracy} is the mean block score of the last tenth of the common tasks, and the \emph{late rare-content accuracy} is the mean block score of the 25 Kuzushiji tasks among tasks 1,501--2,000; compute and storage are averaged over the common-content window.
Three rules decide a verdict: (a)~\emph{not worse on the common content}, the lower end of the interval of the difference to the ordinary network is above $-0.5$ points; (b)~\emph{smaller}, the upper end of the interval of the compute is at most 75\% of the ordinary network; (c)~\emph{rare content kept}, the lower end of the interval of the late rare-content difference is above $-1.0$ point.
Rules (a) and (b) were fixed before the method was run on this stream.
Rule (c) was added after the first results with death and before dormancy was run; dormancy is judged on all three.

\section{Deployment: a network that shrinks while it learns}\label{sec:deployment}

\begin{figure}[t]
  \centering
  \begin{tikzpicture}[x=0.0098\linewidth,y=1cm,font=\scriptsize]
  \fill[cOrdFill] (0,0) rectangle (25,0.42);
  \fill[cGrowFill] (25,0) rectangle (100,0.42);
  \draw[cInkTwo, line width=0.4pt] (0,0) rectangle (100,0.42);
  \draw[cInkTwo, line width=0.4pt] (25,0) -- (25,0.42);
  \foreach \k in {0,...,74} { \draw[cDorm, line width=0.5pt] (25.5+\k,0.04) -- (25.5+\k,0.38); }
  \node[text=cInk, anchor=center, fill=cOrdFill, inner sep=1pt] at (12.5,0.21) {general phase};
  \node[text=cInk, anchor=center, fill=cGrowFill, inner sep=1.5pt] at (62.5,0.21) {deployment phase};
  \node[text=cInkTwo, anchor=north, align=center] at (12.5,-0.06) {digits, letters, Kuzushiji,\\clothing, second sensor};
  \node[text=cInkTwo, anchor=north, align=center] at (62.5,-0.06) {common: digits and letters \quad rare: Kuzushiji, one task in 20 (ticks)\\gone: clothing and the second sensor};
  \draw[cGrow, -{Stealth[length=4pt]}, line width=0.9pt] (25,0.95) -- (25,0.46);
  \node[text=cInk, anchor=south] at (25,0.93) {Growth Path switched on (task 500)};
  \node[text=cInkTwo, anchor=south west] at (0,0.44) {task 1};
  \node[text=cInkTwo, anchor=south east] at (100,0.44) {task 2,000};
\end{tikzpicture}\\[4pt]
  \begin{tikzpicture}
\begin{groupplot}[dpn,
  group style={group size=3 by 2, horizontal sep=1.15cm, vertical sep=0.95cm},
  width=0.345\linewidth, height=3.9cm,
  xmin=500, xmax=2000, xtick={500,1000,1500,2000},
  scaled x ticks=false, x tick label style={/pgf/number format/1000 sep={,}}]
\nextgroupplot[title={units per layer}, ymin=0, ymax=320, ytick={0,100,200,300}, ylabel={\textbf{Adam}}]
  \addplot[cOrd, line width=1.1pt, mark=none] coordinates {(500,300) (2000,300)};
  \addplot[cGrow, mark=none] table[x=task,y=b100m] {fig/data/time_size_adam.dat};
  \addplot[cGrow, mark=none, dash pattern=on 3pt off 1.5pt] table[x=task,y=b050m] {fig/data/time_size_adam.dat};
  \node[font=\scriptsize, anchor=north, text=cInk] at (axis cs:1150,296) {ordinary};
  \node[font=\scriptsize, anchor=south east, text=cInk] at (axis cs:2000,214) {$B=100$};
  \node[font=\scriptsize, anchor=south east, text=cInk] at (axis cs:2000,103) {$B=50$};
\nextgroupplot[title={common content (\%)}, ymin=93.4, ymax=95.8, ytick={93.5,94,94.5,95,95.5}]
  \addplot[cOrd, mark=none] table[x=task,y=ordinary] {fig/data/time_common_adam.dat};
  \addplot[cGrow, mark=none] table[x=task,y=gp100] {fig/data/time_common_adam.dat};
  \addplot[cGrow, mark=none, dash pattern=on 3pt off 1.5pt] table[x=task,y=gp050] {fig/data/time_common_adam.dat};
  \addplot[cRand, mark=none, line width=0.9pt, dash pattern=on 1pt off 1.5pt] table[x=task,y=replay100] {fig/data/time_common_adam.dat};
\nextgroupplot[title={rare content (\%)}, ymin=70, ymax=83, ytick={70,74,78,82}]
  \addplot[cOrd, mark=none] table[x=task,y=ordinary] {fig/data/time_rare_adam.dat};
  \addplot[cGrow, mark=none] table[x=task,y=gp100] {fig/data/time_rare_adam.dat};
  \addplot[cGrow, mark=none, dash pattern=on 3pt off 1.5pt] table[x=task,y=gp050] {fig/data/time_rare_adam.dat};
  \addplot[cRand, mark=none, line width=0.9pt, dash pattern=on 1pt off 1.5pt] table[x=task,y=replay100] {fig/data/time_rare_adam.dat};
\nextgroupplot[ymin=0, ymax=320, ytick={0,100,200,300}, ylabel={\textbf{SGD}}, xlabel={task}]
  \addplot[cOrd, line width=1.1pt, mark=none] coordinates {(500,300) (2000,300)};
  \addplot[cGrow, mark=none] table[x=task,y=b100m] {fig/data/time_size_sgd.dat};
  \addplot[cGrow, mark=none, dash pattern=on 3pt off 1.5pt] table[x=task,y=b050m] {fig/data/time_size_sgd.dat};
\nextgroupplot[ymin=94.5, ymax=96.9, ytick={94.5,95,95.5,96,96.5}, xlabel={task},
  legend to name=deploytimelegend, legend columns=4, legend style={/tikz/every even column/.append style={column sep=8pt}}]
  \addplot[cOrd, mark=none] table[x=task,y=ordinary] {fig/data/time_common_sgd.dat};
  \addlegendentry{ordinary network}
  \addplot[cGrow, mark=none] table[x=task,y=gp100] {fig/data/time_common_sgd.dat};
  \addlegendentry{Growth Path, $B=100$}
  \addplot[cGrow, mark=none, dash pattern=on 3pt off 1.5pt] table[x=task,y=gp050] {fig/data/time_common_sgd.dat};
  \addlegendentry{Growth Path, $B=50$}
  \addplot[cRand, mark=none, line width=0.9pt, dash pattern=on 1pt off 1.5pt] table[x=task,y=replay100] {fig/data/time_common_sgd.dat};
  \addlegendentry{random turnover, $B=100$}
\nextgroupplot[ymin=79, ymax=89.5, ytick={80,82,84,86,88}, xlabel={task}]
  \addplot[cOrd, mark=none] table[x=task,y=ordinary] {fig/data/time_rare_sgd.dat};
  \addplot[cGrow, mark=none] table[x=task,y=gp100] {fig/data/time_rare_sgd.dat};
  \addplot[cGrow, mark=none, dash pattern=on 3pt off 1.5pt] table[x=task,y=gp050] {fig/data/time_rare_sgd.dat};
  \addplot[cRand, mark=none, line width=0.9pt, dash pattern=on 1pt off 1.5pt] table[x=task,y=replay100] {fig/data/time_rare_sgd.dat};
\end{groupplot}
\end{tikzpicture}\par\vspace{2pt}
\pgfplotslegendfromname{deploytimelegend}
  \caption{\textbf{Switched on at deployment, the \GP{} shrinks the network while it keeps learning.}
  Top: the deployment stream.
  Bottom: mean number of live units per hidden layer, accuracy on the common content (windows of 100 tasks) and on the rare content (windows of 250 tasks), means of five seeds.
  The size settles within about 200 tasks at a level set by the budget~$B$.
  With Adam the smaller network is more accurate than the ordinary network on both kinds of content, and a random choice of the same number of units is not.
  With SGD it matches the ordinary network on the common content and falls behind on the rare content.}
  \label{fig:deploy-time}
\end{figure}
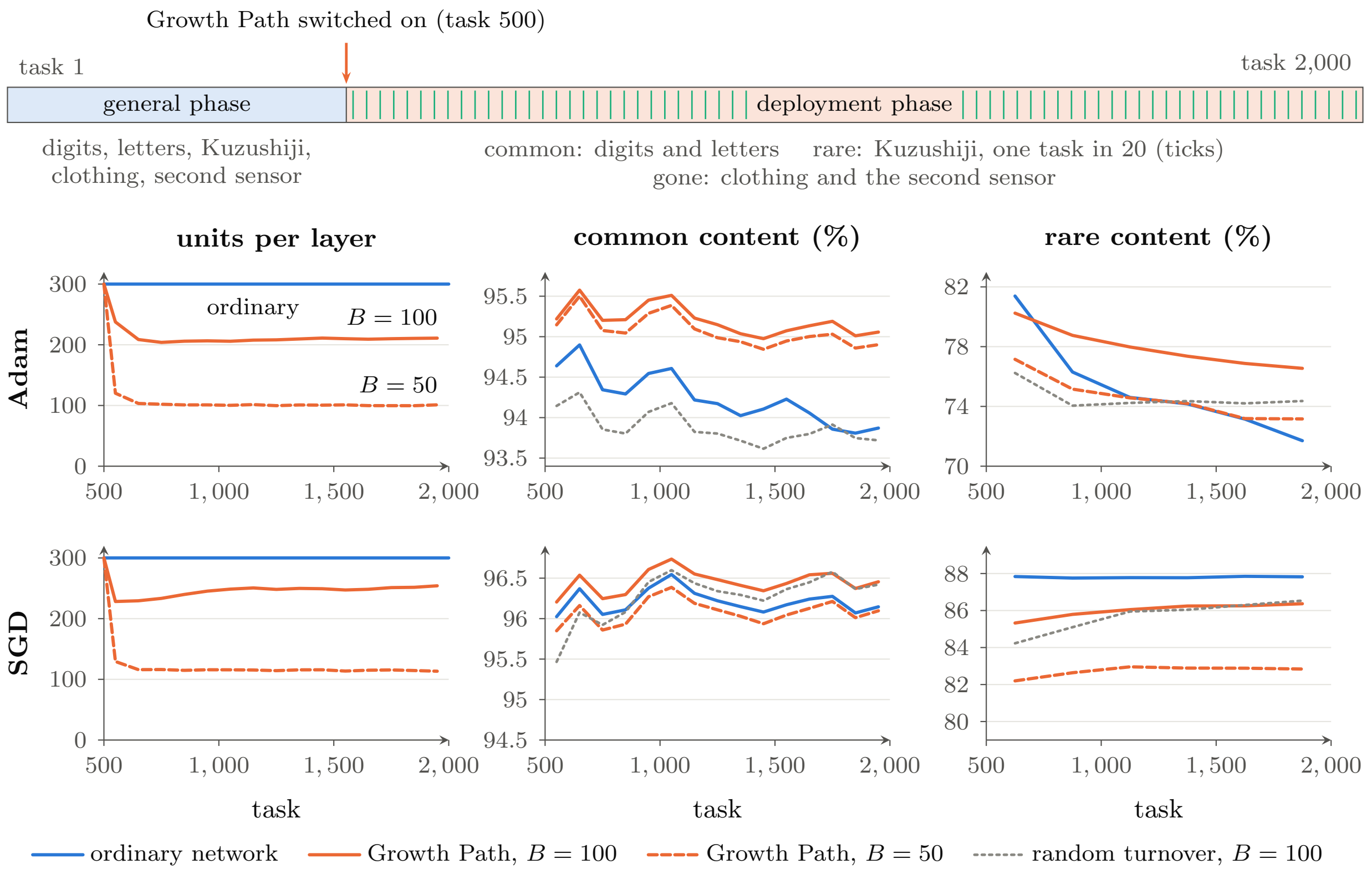

A general model is rarely needed in full where it is used.
A network trained on five kinds of content is deployed on a domain that keeps two of them, shows a third only rarely, and never shows the other two again (Figure~\ref{fig:deploy-time}, top).
An ordinary network carries all of its units through this change.
We switch the \GP{} on at the moment of deployment, in the same pretrained network, and give every layer a budget smaller than its width.
Nothing else is changed and nothing is tuned on this stream (Section~\ref{sec:seeds}).

\subsection{The ordinary network, left to itself}\label{sec:deploy-ordinary}

The two optimisers give very different baselines, and every result below has to be read against its own.
With Adam the ordinary network reaches 93.9\% on the common content, and it loses the rare content by itself: Kuzushiji accuracy falls from 87.3\% at the end of the general phase to 72.4\% late in deployment.
Most of its units have stopped working: late in deployment 30\%, 94\% and 92\% of the units of layers 1, 2 and 3 are inactive on held-out images.
With SGD the ordinary network reaches 96.1\%, keeps the rare content (88.7\% to 87.8\%), and has 9\%, 60\% and 35\% inactive units.
So under Adam there is much to gain and little to lose, and under SGD the ordinary network is a strong baseline that loses nothing.

\subsection{Smaller and, with Adam, better}\label{sec:deploy-main}

Figure~\ref{fig:deploy-time} shows what happens after the switch-on, and Table~\ref{tab:deploy} gives the verdicts.
Within about 200 tasks the layers shrink to a size set by the budget, roughly two units per unit of budget (about 200 units per layer for $B=100$ and 100 for $B=50$), and then stay there.
The network keeps learning throughout.

\emph{Adam.}
With $B=100$ the network ends 1.16 points above the ordinary network on the common content (95.07\% against 93.92\%) at 60\% of its compute.
With $B=50$ it is 1.01 points above at 24\% of the compute.
Both rules fixed in advance hold, with all five seeds on the same side.
The rare content is not lost either: $+4.3$ points at $B=100$ and no measurable difference at $B=50$.
The share of inactive units falls to 14\%, 6\% and 44\% in the three layers at $B=100$: the units that remain are units that work.

\emph{SGD.}
With $B=50$ the network ties the ordinary network on the common content ($-0.05$ points) at 31\% of the compute, so rules (a) and (b) hold.
It pays with the rare content, which ends 5.0 points lower.
With $B=100$ it is 0.31 points better on the common content, but the compute falls by 23\% only, short of the 25\% that rule (b) asks for, and the rare content is 1.5 points lower.

\emph{No reduced budget, no shrinking.}
With $B=300$, the width of the pretrained network, every run of both optimisers grew until a layer reached the storage limit of 768 units and was stopped.
The budget has to be chosen below the width for the network to shrink; Section~\ref{sec:budget} explains why.

\begin{table}[t]
  \centering
  \small
  \begin{tabular}{@{}lrllccc@{}}
\toprule
 & Compute & Common content & Late rare content & \multicolumn{3}{c}{Rules} \\
Network & (\%) & (points) & (points) & (a) & (b) & (c) \\
\midrule
\multicolumn{7}{@{}l}{\textbf{Adam}} \\
Ordinary network & 100 & 93.92 & 72.41 & & & \\
\addlinespace[2pt]
Growth Path, $B=100$ & 60 & $+1.16$ {\scriptsize[$0.76$, $1.55$]} & $+4.3$ {\scriptsize[$1.1$, $7.5$]} & \checkmark & \checkmark & \checkmark \\
Growth Path, $B=50$ & 24 & $+1.01$ {\scriptsize[$0.69$, $1.33$]} & $+0.8$ {\scriptsize[$-2.7$, $4.3$]} & \checkmark & \checkmark & $\times$ \\
Growth Path, $B=25$ & 12 & $+0.54$ {\scriptsize[$0.21$, $0.86$]} & $-3.5$ {\scriptsize[$-7.0$, $-0.1$]} & \checkmark & \checkmark & $\times$ \\
\addlinespace[2pt]
\quad with dormancy, $B=100$ & 71 & $+1.20$ {\scriptsize[$0.87$, $1.53$]} & $+8.1$ {\scriptsize[$5.1$, $11.2$]} & \checkmark & \checkmark & \checkmark \\
\quad with dormancy, $B=50$ & 31 & $+1.04$ {\scriptsize[$0.63$, $1.44$]} & $+5.4$ {\scriptsize[$2.2$, $8.5$]} & \checkmark & \checkmark & \checkmark \\
\quad with dormancy, $B=35$ & 22 & $+0.84$ {\scriptsize[$0.55$, $1.13$]} & $+3.8$ {\scriptsize[$1.3$, $6.4$]} & \checkmark & \checkmark & \checkmark \\
\midrule
\multicolumn{7}{@{}l}{\textbf{SGD}} \\
Ordinary network & 100 & 96.15 & 87.84 & & & \\
\addlinespace[2pt]
Growth Path, $B=100$ & 77 & $+0.31$ {\scriptsize[$0.16$, $0.45$]} & $-1.5$ {\scriptsize[$-2.0$, $-1.0$]} & \checkmark & $\times$ & $\times$ \\
Growth Path, $B=50$ & 31 & $-0.05$ {\scriptsize[$-0.17$, $0.07$]} & $-5.0$ {\scriptsize[$-5.5$, $-4.4$]} & \checkmark & \checkmark & $\times$ \\
Growth Path, $B=25$ & 15 & $-0.85$ {\scriptsize[$-0.92$, $-0.78$]} & $-11.7$ {\scriptsize[$-12.8$, $-10.5$]} & $\times$ & \checkmark & $\times$ \\
\addlinespace[2pt]
\quad with dormancy, $B=100$ & 85 & $+0.31$ {\scriptsize[$0.20$, $0.42$]} & $-0.3$ {\scriptsize[$-0.7$, $0.1$]} & \checkmark & $\times$ & \checkmark \\
\quad with dormancy, $B=50$ & 39 & $+0.04$ {\scriptsize[$-0.09$, $0.17$]} & $-2.8$ {\scriptsize[$-3.4$, $-2.2$]} & \checkmark & \checkmark & $\times$ \\
\quad with dormancy, $B=35$ & 27 & $-0.17$ {\scriptsize[$-0.31$, $-0.04$]} & $-5.5$ {\scriptsize[$-5.8$, $-5.1$]} & \checkmark & \checkmark & $\times$ \\
\bottomrule
\end{tabular}

  \caption{\textbf{Deployment: differences to the ordinary network} (seeds 15--19, paired by seed, 95\% intervals).
  Compute is the counted cost as a percentage of the ordinary network.
  The first row of each block gives the accuracy of the ordinary network in percent.
  Rules, fixed in advance (Section~\ref{sec:verdicts}): (a) common content not more than 0.5 points worse; (b) compute at most 75\%; (c) late rare content not more than 1.0 point worse.
  Each is judged on the end of the interval, so a positive mean with a wide interval can fail rule~(c).
  The \GP{} without dormancy is judged on (a) and (b); rule (c) was added for dormancy and is shown for all rows.
  Appendix~\ref{app:tables} lists all six budgets.}
  \label{tab:deploy}
\end{table}

\subsection{One knob for compute}\label{sec:deploy-curve}

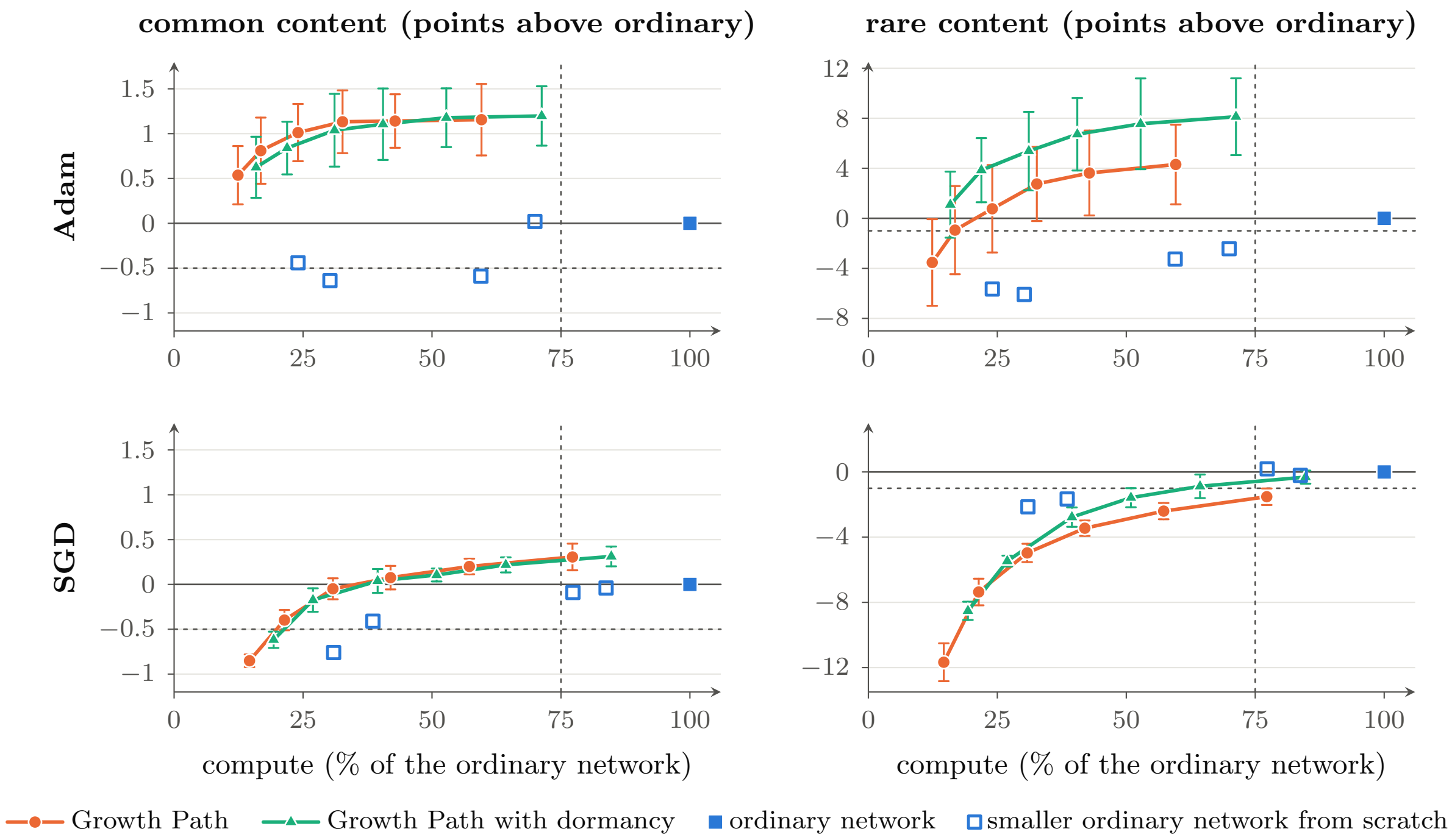
\begin{figure}[t]
  \centering
  \begin{tikzpicture}
\begin{groupplot}[dpn,
  group style={group size=2 by 2, horizontal sep=1.6cm, vertical sep=1.0cm},
  width=0.455\linewidth, height=4.5cm,
  xmin=0, xmax=106, xtick={0,25,50,75,100},
  error bars/y dir=both, error bars/y explicit,
  error bars/error bar style={line width=0.6pt}, error bars/error mark options={rotate=90, mark size=1.5pt, line width=0.6pt}]
\nextgroupplot[title={common content (points above ordinary)}, ymin=-1.2, ymax=1.8, ytick={-1,-0.5,0,0.5,1,1.5}, ylabel={\textbf{Adam}}]
  \addplot[cInkTwo, line width=0.5pt, mark=none, forget plot, error bars/.cd, y dir=none] coordinates {(0,0) (106,0)};
  \addplot[cInkTwo, line width=0.5pt, dash pattern=on 1pt off 2pt, mark=none, forget plot, error bars/.cd, y dir=none] coordinates {(75,-1.2) (75,1.8)};
  \addplot[cInkTwo, line width=0.5pt, dash pattern=on 1pt off 2pt, mark=none, forget plot, error bars/.cd, y dir=none] coordinates {(0,-0.5) (106,-0.5)};
  \addplot[growline] table[x=rel,y=dc,y error minus=dcm,y error plus=dcp] {fig/data/cost_adam_death.dat};
  \addplot[dormline] table[x=rel,y=dc,y error minus=dcm,y error plus=dcp] {fig/data/cost_adam_dormancy.dat};
  \addplot[scratchpts, error bars/.cd, y dir=none] table[x=rel,y=dc] {fig/data/scratch_adam.dat};
  \addplot[ordline, only marks, mark size=2.3pt, error bars/.cd, y dir=none] coordinates {(100,0)};
\nextgroupplot[title={rare content (points above ordinary)}, ymin=-9, ymax=12.5, ytick={-8,-4,0,4,8,12}]
  \addplot[cInkTwo, line width=0.5pt, mark=none, forget plot, error bars/.cd, y dir=none] coordinates {(0,0) (106,0)};
  \addplot[cInkTwo, line width=0.5pt, dash pattern=on 1pt off 2pt, mark=none, forget plot, error bars/.cd, y dir=none] coordinates {(75,-9) (75,12.5)};
  \addplot[cInkTwo, line width=0.5pt, dash pattern=on 1pt off 2pt, mark=none, forget plot, error bars/.cd, y dir=none] coordinates {(0,-1) (106,-1)};
  \addplot[growline] table[x=rel,y=dr,y error minus=drm,y error plus=drp] {fig/data/cost_adam_death.dat};
  \addplot[dormline] table[x=rel,y=dr,y error minus=drm,y error plus=drp] {fig/data/cost_adam_dormancy.dat};
  \addplot[scratchpts, error bars/.cd, y dir=none] table[x=rel,y=dr] {fig/data/scratch_adam.dat};
  \addplot[ordline, only marks, mark size=2.3pt, error bars/.cd, y dir=none] coordinates {(100,0)};
\nextgroupplot[ymin=-1.2, ymax=1.8, ytick={-1,-0.5,0,0.5,1,1.5}, ylabel={\textbf{SGD}}, xlabel={compute (\% of the ordinary network)},
  legend to name=costlegend, legend columns=4, legend style={/tikz/every even column/.append style={column sep=8pt}}]
  \addplot[cInkTwo, line width=0.5pt, mark=none, forget plot, error bars/.cd, y dir=none] coordinates {(0,0) (106,0)};
  \addplot[cInkTwo, line width=0.5pt, dash pattern=on 1pt off 2pt, mark=none, forget plot, error bars/.cd, y dir=none] coordinates {(75,-1.2) (75,1.8)};
  \addplot[cInkTwo, line width=0.5pt, dash pattern=on 1pt off 2pt, mark=none, forget plot, error bars/.cd, y dir=none] coordinates {(0,-0.5) (106,-0.5)};
  \addplot[growline] table[x=rel,y=dc,y error minus=dcm,y error plus=dcp] {fig/data/cost_sgd_death.dat};
  \addlegendentry{Growth Path}
  \addplot[dormline] table[x=rel,y=dc,y error minus=dcm,y error plus=dcp] {fig/data/cost_sgd_dormancy.dat};
  \addlegendentry{Growth Path with dormancy}
  \addplot[ordline, only marks, mark size=2.3pt, error bars/.cd, y dir=none] coordinates {(100,0)};
  \addlegendentry{ordinary network}
  \addplot[scratchpts, error bars/.cd, y dir=none] table[x=rel,y=dc] {fig/data/scratch_sgd.dat};
  \addlegendentry{smaller ordinary network from scratch}
\nextgroupplot[ymin=-13.5, ymax=3, ytick={-12,-8,-4,0}, xlabel={compute (\% of the ordinary network)}]
  \addplot[cInkTwo, line width=0.5pt, mark=none, forget plot, error bars/.cd, y dir=none] coordinates {(0,0) (106,0)};
  \addplot[cInkTwo, line width=0.5pt, dash pattern=on 1pt off 2pt, mark=none, forget plot, error bars/.cd, y dir=none] coordinates {(75,-13.5) (75,3)};
  \addplot[cInkTwo, line width=0.5pt, dash pattern=on 1pt off 2pt, mark=none, forget plot, error bars/.cd, y dir=none] coordinates {(0,-1) (106,-1)};
  \addplot[growline] table[x=rel,y=dr,y error minus=drm,y error plus=drp] {fig/data/cost_sgd_death.dat};
  \addplot[dormline] table[x=rel,y=dr,y error minus=drm,y error plus=drp] {fig/data/cost_sgd_dormancy.dat};
  \addplot[scratchpts, error bars/.cd, y dir=none] table[x=rel,y=dr] {fig/data/scratch_sgd.dat};
  \addplot[ordline, only marks, mark size=2.3pt, error bars/.cd, y dir=none] coordinates {(100,0)};
\end{groupplot}
\end{tikzpicture}\par\vspace{2pt}
\pgfplotslegendfromname{costlegend}
  \caption{\textbf{The budget trades compute against accuracy along a smooth curve.}
  Difference to the ordinary network against compute, for budgets $B=25$, 35, 50, 65, 80 and 100 (left to right on each curve); bars are 95\% intervals of the paired differences.
  Dotted lines mark the rules of Table~\ref{tab:deploy}.
  With Adam the \GP{} is above the ordinary network on the common content at every budget, down to 12\% of the compute, and dormancy keeps the rare content as well.
  With SGD the common content holds down to about 30\% of the compute and the rare content is lost progressively.
  Open squares: ordinary networks of the sizes reached at $B=50$ and $B=100$, trained from scratch on the whole stream.}
  \label{fig:cost}
\end{figure}

The budget is the only number we changed to obtain Figure~\ref{fig:cost}.
Six budgets from 25 to 100 give six networks whose compute runs from 12\% to 60\% of the ordinary network under Adam and from 15\% to 77\% under SGD, in order and without any other setting.
With Adam, rules (a) and (b) hold at every budget.
At $B=25$ the network works with 59, 46 and 45 units in its three layers, 12\% of the compute, and is still 0.54 points above the 300-wide ordinary network on the common content.
With SGD the common content is kept down to $B=50$ and then gives way: $-0.40$ points at $B=35$ and $-0.85$ at $B=25$.

\subsection{It matters which units go}\label{sec:deploy-which}

A smaller network with fresh units might be better for reasons that have nothing to do with the nutrient.
Two controls test this.

\emph{Random turnover} repeats each run with the same number of deaths and divisions at every check, drawing the units at random.
Under Adam it ends no better than the ordinary network (93.8\% and 93.9\% against 93.9\%), and the \GP{} is 1.30 points above it at $B=100$ and 1.04 points at $B=50$ (Figure~\ref{fig:deploy-time}).
The gain under Adam therefore comes from which units are replaced, not from replacing units.
Under SGD the two are close: $+0.02$ points at $B=100$ and $+0.27$ at $B=50$.

An \emph{ordinary network of the same final size, trained from scratch} on the whole stream, is 1.74 and 1.45 points below the \GP{} on the common content under Adam, and 0.40 and 0.71 points below it under SGD.
Size alone does not explain the result either.

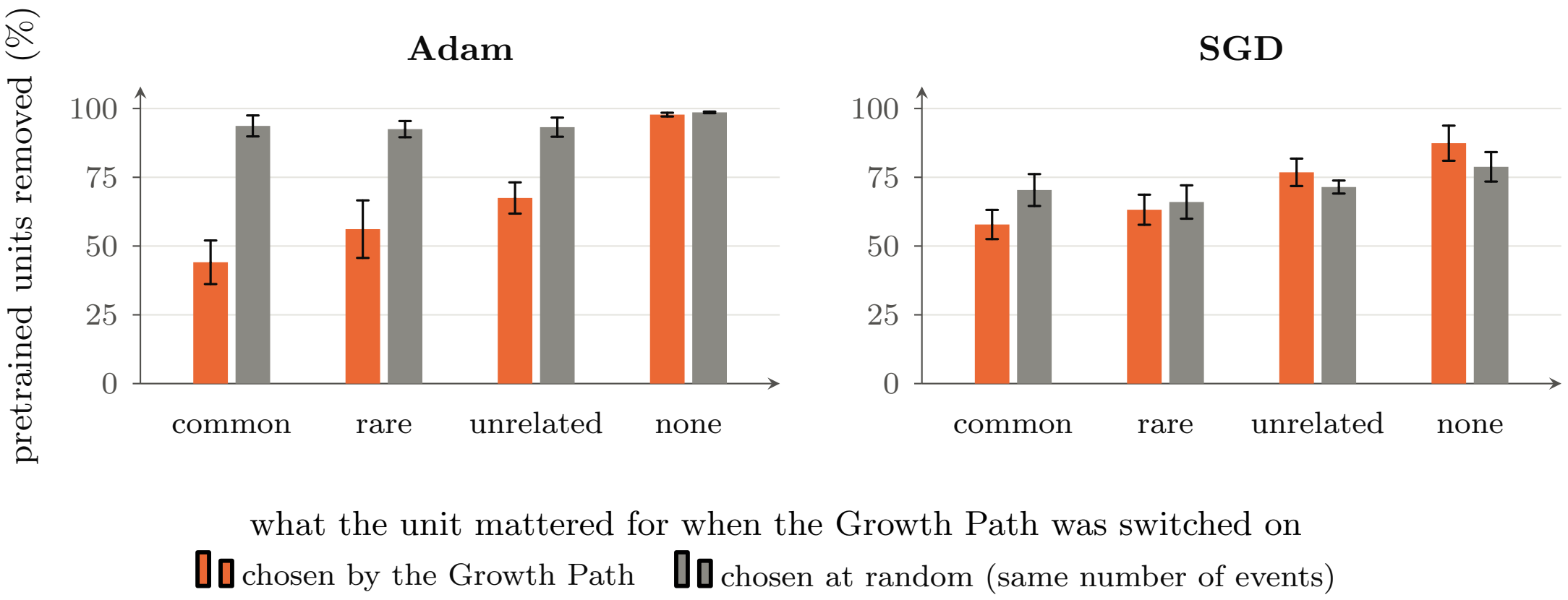
\begin{figure}[t]
  \centering
  \begin{tikzpicture}
\begin{groupplot}[dpn,
  group style={group size=2 by 1, horizontal sep=1.3cm},
  width=0.45\linewidth, height=4.3cm,
  ybar, /pgf/bar width=9pt, ymin=0, ymax=108, ytick={0,25,50,75,100},
  xmin=-0.6, xmax=3.6, xtick={0,1,2,3}, xticklabels={common,rare,unrelated,none},
  x tick label style={font=\scriptsize, text=cInk, text height=1.5ex, text depth=0.4ex}, xtick style={draw=none},
  enlarge x limits=false,
  error bars/y dir=both, error bars/y explicit,
  error bars/error bar style={line width=0.6pt, cInk}, error bars/error mark options={rotate=90, mark size=1.5pt, line width=0.6pt, cInk}]
\nextgroupplot[title={Adam}, ylabel={pretrained units removed (\%)},
  legend to name=whodieslegend, legend columns=2, legend style={/tikz/every even column/.append style={column sep=8pt}}]
  \addplot[draw=none, fill=cGrow, bar shift=-5.5pt] table[x=x,y=rule,y error minus=rulem,y error plus=rulep] {fig/data/whodies_adam_b100.dat};
  \addlegendentry{chosen by the Growth Path}
  \addplot[draw=none, fill=cRand, bar shift=5.5pt] table[x=x,y=rand,y error minus=randm,y error plus=randp] {fig/data/whodies_adam_b100.dat};
  \addlegendentry{chosen at random (same number of events)}
\nextgroupplot[title={SGD}]
  \addplot[draw=none, fill=cGrow, bar shift=-5.5pt] table[x=x,y=rule,y error minus=rulem,y error plus=rulep] {fig/data/whodies_sgd_b100.dat};
  \addplot[draw=none, fill=cRand, bar shift=5.5pt] table[x=x,y=rand,y error minus=randm,y error plus=randp] {fig/data/whodies_sgd_b100.dat};
\end{groupplot}
\end{tikzpicture}\par\vspace{1pt}
{\footnotesize what the unit mattered for when the Growth Path was switched on}\par\vspace{2pt}
\pgfplotslegendfromname{whodieslegend}
  \caption{\textbf{The rule removes what the domain no longer needs, without ever seeing a label.}
  Share of the pretrained units removed in the first 100 tasks after switch-on ($B=100$), by the content a unit mattered for when the \GP{} was switched on (the content whose loss rises most when the unit is silenced; ``none'': silencing it changes nothing).
  Bars are means of five seeds with 95\% intervals.
  A random choice of the same number of units makes no such difference.}
  \label{fig:whodies}
\end{figure}

Figure~\ref{fig:whodies} looks at the units themselves.
Before the switch-on we classified every pretrained unit by silencing it: a unit matters for the common content, the rare content or the content that is gone, according to where the loss rises most, or for none if nothing changes.
The \GP{} sees none of this; it reads activations and weights.
Yet after 100 tasks it has removed 98\% of the units that mattered for nothing, 67\% of those that served content that is gone, and 44\% of those that served the common content (Adam, $B=100$).
Units for content that is gone die 23 points more often than units for the common content under Adam and 19 points more often under SGD, and with a random choice the difference cannot be told from zero.
The selection is real but far from clean: by the end of the stream about 70\% of the units that mattered for the common content have been replaced as well.
The network is rebuilt, not just trimmed (Section~\ref{sec:analysis}).

\subsection{Dormancy keeps what is rarely needed}\label{sec:deploy-dormancy}

Death is final, and the rare content shows its price: a unit used in one task out of twenty can starve in the nineteen in between.
With dormancy (Section~\ref{sec:dormancy}) a starving pretrained unit is set aside with everything it has learned, and returns when the nutrient it would receive reaches the mean of its layer.

\begin{figure}[t]
  \centering
  \begin{minipage}[b]{0.49\linewidth}\centering
\begin{tikzpicture}
\begin{axis}[dpn, width=\linewidth, height=4.4cm,
  ybar, /pgf/bar width=9pt, ymin=0, ymax=185, ytick={0,50,100,150},
  xmin=-0.6, xmax=3.6, xtick={0,1,2,3},
  xticklabels={{Adam\\$B{=}100$},{Adam\\$B{=}50$},{SGD\\$B{=}100$},{SGD\\$B{=}50$}},
  x tick label style={font=\scriptsize, text=cInk, align=center}, xtick style={draw=none},
  ylabel={wakes per task},
  legend style={at={(0.98,0.98)}, anchor=north east},
  error bars/y dir=both, error bars/y explicit,
  error bars/error bar style={line width=0.6pt, cInk}, error bars/error mark options={rotate=90, mark size=1.5pt, line width=0.6pt, cInk}]
  \addplot[draw=none, fill=cDorm, bar shift=-5.5pt] coordinates {(0,151.4) +- (0,11.2) (1,121.6) +- (0,14.7) (2,55.8) +- (0,7.4) (3,27.5) +- (0,4.2)};
  \addlegendentry{rare-content task}
  \addplot[draw=none, fill=cRand, bar shift=5.5pt] coordinates {(0,7.6) +- (0,1.2) (1,5.7) +- (0,0.9) (2,4.2) +- (0,0.8) (3,2.8) +- (0,0.6)};
  \addlegendentry{common-content task}
\end{axis}
\end{tikzpicture}\\[-2pt]
{\footnotesize (a)}
\end{minipage}\hfill
\begin{minipage}[b]{0.46\linewidth}\centering
\begin{tikzpicture}
\begin{axis}[dpn, width=\linewidth, height=4.4cm,
  ybar stacked, /pgf/bar width=14pt, ymin=0, ymax=100, ytick={0,25,50,75,100},
  xmin=-0.6, xmax=3.6, xtick={0,1,2,3}, xticklabels={common,rare,unrelated,none},
  x tick label style={font=\scriptsize, text=cInk, text height=1.5ex, text depth=0.4ex}, xtick style={draw=none},
  ylabel={founder units (\%)},
  legend style={at={(0.5,1.04)}, anchor=south, legend columns=3, /tikz/every even column/.append style={column sep=5pt}}]
  \addplot[draw=white, line width=0.8pt, fill=cGrow] coordinates {(0,51) (1,51) (2,36) (3,5)};
  \addlegendentry{live}
  \addplot[draw=white, line width=0.8pt, fill=cDorm] coordinates {(0,46) (1,44) (2,54) (3,71)};
  \addlegendentry{dormant, woke before}
  \addplot[draw=white, line width=0.8pt, fill=cRand!55] coordinates {(0,3) (1,4) (2,11) (3,24)};
  \addlegendentry{dormant, never woke}
\end{axis}
\end{tikzpicture}\\[-2pt]
{\footnotesize (b)}
\end{minipage}
  \caption{\textbf{Dormant units wake when the content they serve returns.}
  (a) Wakes per task on rare-content tasks and on common-content tasks (means of five seeds, 95\% intervals).
  (b) The pretrained units at the end of the stream, by the content they mattered for at switch-on (Adam, $B=100$).}
  \label{fig:dormancy}
\end{figure}
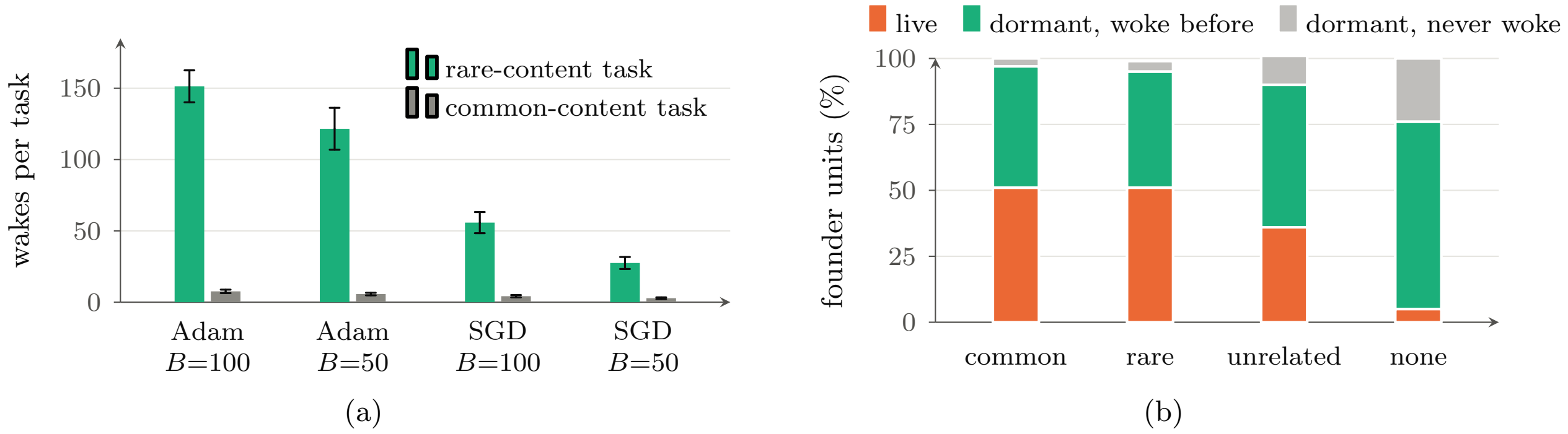

Dormancy keeps more of the rare content than death in all four settings that we ran with both: $+3.8$ and $+4.6$ points under Adam ($B=100$ and 50) and $+1.2$ and $+2.2$ points under SGD, with every interval above zero.
The common content is unchanged.
The price is the wake test and the units that are awake at any time, 7 to 12 percentage points of compute.
Under Adam all three rules now hold at every budget from 35 to 100 (Table~\ref{tab:deploy}, Figure~\ref{fig:cost}): at $B=35$ the network uses 22\% of the compute, is 0.84 points better on the common content and 3.8 points better on the rare content than the ordinary network.
At equal compute, dormancy keeps about 2.5 to 3.6 points more rare content than death along the whole curve.

The wakes follow the content (Figure~\ref{fig:dormancy}a).
A rare-content task triggers 122 to 151 wakes under Adam and 28 to 56 under SGD; a common-content task triggers 3 to 8.
No part of the rule knows which task is running.
Figure~\ref{fig:dormancy}b shows where the pretrained units end up: about half of those that mattered for the common or the rare content are live at the end, and almost all of the others have been awake at some time; the units that mattered for nothing are mostly dormant.

Dormancy saves compute, not storage.
Every pretrained unit is kept, so the stored network is 1.1 to 1.8 times the original.

\subsection{With SGD: a trade-off, stated plainly}\label{sec:deploy-sgd}

Under SGD the same curves describe a trade-off and not a free gain (Figure~\ref{fig:cost}, bottom).
The common content is kept down to about 30\% of the compute.
The rare content falls behind the ordinary network at every budget, from 1.5 points at $B=100$ to 11.7 points at $B=25$; with dormancy from 0.3 to 8.5 points.
At equal compute, dormancy is about one point better than death from about 40\% of the compute upwards and less than half a point better below.
No budget meets all three rules: at $B=100$ the rare content ties the ordinary network but the compute is only 15\% lower, and from $B=80$ down the rare content is lost.

One comparison goes against the method and belongs here.
An ordinary network of the reduced size, trained from scratch on the whole stream with SGD, loses \emph{less} rare content than the \GP{} at the same compute: 2.1 against 5.0 points at 31\% of the compute, and 1.7 against 2.8 points (with dormancy) at 39\%.
It loses more of the common content (0.76 and 0.41 points), and it presupposes that the final size is known in advance and that the general phase can be repeated at that size, which is not the situation of deployment.
But under SGD a network that never had the extra units keeps the rare content better than one that sheds them.

\section{Continual learning on long task streams}\label{sec:continual}

Deployment starts from a network that already exists.
Here the network is trained from scratch on a stream that keeps changing, the \GP{} is on from the first update, and the budget equals the starting width.
The network chooses its own size, so the ordinary network is run at the sizes the method reaches (Section~\ref{sec:baseline}).
We use two streams that differ in what old knowledge is worth.

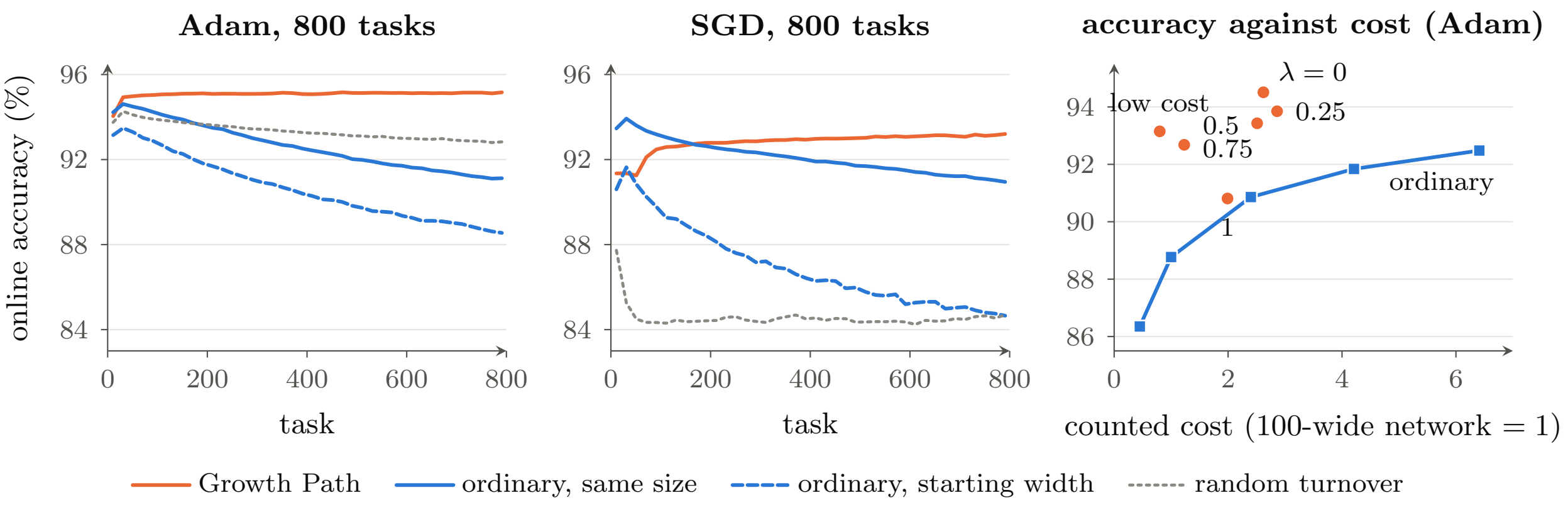
\begin{figure}[t]
  \centering
  \begin{tikzpicture}
\begin{groupplot}[dpn,
  group style={group size=3 by 1, horizontal sep=1.1cm},
  width=0.35\linewidth, height=4.6cm]
\nextgroupplot[title={Adam, 800 tasks}, xmin=0, xmax=800, xtick={0,200,400,600,800}, ymin=83, ymax=96.5, ytick={84,88,92,96},
  xlabel={task}, ylabel={online accuracy (\%)},
  legend to name=continuallegend, legend columns=4, legend style={/tikz/every even column/.append style={column sep=8pt}}]
  \addplot[cGrow, mark=none] table[x=task,y=growth] {fig/data/long_adam_n300.dat};
  \addlegendentry{Growth Path}
  \addplot[cOrd, mark=none] table[x=task,y=peak] {fig/data/long_adam_n300.dat};
  \addlegendentry{ordinary, same size}
  \addplot[cOrd, mark=none, dash pattern=on 3pt off 1.5pt] table[x=task,y=start] {fig/data/long_adam_n300.dat};
  \addlegendentry{ordinary, starting width}
  \addplot[cRand, mark=none, line width=0.9pt, dash pattern=on 1pt off 1.5pt] table[x=task,y=random] {fig/data/long_adam_n300.dat};
  \addlegendentry{random turnover}
\nextgroupplot[title={SGD, 800 tasks}, xmin=0, xmax=800, xtick={0,200,400,600,800}, ymin=83, ymax=96.5, ytick={84,88,92,96}, xlabel={task}]
  \addplot[cGrow, mark=none] table[x=task,y=growth] {fig/data/long_sgd_n100.dat};
  \addplot[cOrd, mark=none] table[x=task,y=peak] {fig/data/long_sgd_n100.dat};
  \addplot[cOrd, mark=none, dash pattern=on 3pt off 1.5pt] table[x=task,y=start] {fig/data/long_sgd_n100.dat};
  \addplot[cRand, mark=none, line width=0.9pt, dash pattern=on 1pt off 1.5pt] table[x=task,y=random] {fig/data/long_sgd_n100.dat};
\nextgroupplot[title={accuracy against cost (Adam)}, xmin=0, xmax=7, xtick={0,2,4,6}, ymin=85.5, ymax=95.5, ytick={86,88,90,92,94},
  xlabel={counted cost (100-wide network $=1$)}]
  \addplot[ordline] coordinates {(0.45,86.35) (1.00,88.77) (2.40,90.86) (4.21,91.84) (6.41,92.48)};
  \addplot[growline, only marks] coordinates {(2.62,94.51) (2.86,93.85) (2.51,93.43) (1.23,92.68) (1.99,90.81) (0.80,93.15)};
  \node[font=\scriptsize, text=cInk, anchor=south west] at (axis cs:2.66,94.48) {$\lambda=0$};
  \node[font=\scriptsize, text=cInk, anchor=west] at (axis cs:2.96,93.85) {$0.25$};
  \node[font=\scriptsize, text=cInk, anchor=east] at (axis cs:2.42,93.35) {$0.5$};
  \node[font=\scriptsize, text=cInk, anchor=west] at (axis cs:1.33,92.55) {$0.75$};
  \node[font=\scriptsize, text=cInk, anchor=north] at (axis cs:1.99,90.55) {$1$};
  \node[font=\scriptsize, text=cInk, anchor=south] at (axis cs:0.80,93.40) {low cost};
  \node[font=\scriptsize, text=cInk, anchor=north east] at (axis cs:6.9,92.2) {ordinary};
\end{groupplot}
\end{tikzpicture}\par\vspace{2pt}
\pgfplotslegendfromname{continuallegend}
  \caption{\textbf{On a stream that keeps rewiring its input, a network with the \GP{} keeps learning.}
  Left and middle: online accuracy over 800 pixel-permutation tasks (blocks of 20 tasks, five seeds; $\lambda=0$).
  The ordinary network of the same size is the stronger of our two size-matched baselines (peak size).
  With Adam the \GP{} is ahead throughout; with SGD it starts behind the larger ordinary network, draws level after about 200 tasks and is ahead from then on.
  Right: accuracy against cost after 200 tasks (Adam); each orange point is the setting selected for one inheritance ratio~$\lambda$ (the point marked ``low cost'' is a second setting with $\lambda=0.25$), and the blue line is the ordinary network at widths 50 to 400.}
  \label{fig:continual}
\end{figure}

\subsection{When old knowledge expires: the pixel-permutation stream}\label{sec:permuted}

Every task of this stream permutes the pixels anew, so nothing learned about the input carries over, and an ordinary network gets worse at learning the longer it has been trained.
Figure~\ref{fig:continual} (left) shows 800 tasks under Adam, from a starting width of 300.
The network with the \GP{} rises from 94.0\% to 95.2\% and stays there.
The ordinary network of the starting width falls from 93.1\% to 88.6\%, and the ordinary networks of the sizes the method reaches fall from 93.8\% and 94.2\% to 90.9\% and 91.1\%.
The advantage over the stronger of them grows from $-0.2$ points in the first 20 tasks to $+4.0$ points in the last 20.
Random turnover with the same events also declines (93.8\% to 92.8\%), so that the gap between rule and random choice widens along the whole stream, to $+2.3$ points at the end.
After growing from 900 to 1,488 units around task 100, the network becomes smaller as it goes and ends with 1,230.

Under SGD (middle; starting width 100) the method starts 2.1 points behind the larger ordinary network, draws level after about 200 tasks, and ends 2.3 points ahead, still rising while the ordinary network declines.
Here the network grows instead, from 604 to 1,217 units, and random turnover is far worse than everything else.

\emph{Inheritance.}
On this stream a child does best when it inherits nothing (Figure~\ref{fig:continual}, right).
After 200 tasks under Adam, the accuracy falls from 94.5\% at $\lambda=0$ through 93.4\% at $\lambda=0.5$ to 90.8\% at $\lambda=1$; at $\lambda=0$ the network is 3.2 points above the ordinary network of its final size.
This is as it should be when every task rewires the input: what a parent has learned about its inputs is of no use to the child.
(The SGD runs of this comparison were cut by a deadline and not repeated, so the inheritance curve is for Adam only.)

\emph{Cost.}
A setting with $\lambda=0.25$ settles at 83, 67 and 72 units and reaches 93.1\% at a counted cost of 0.80, less than the 100-wide network it started from.
The ordinary network needs 400 units per layer and eight times the cost (6.41) to reach 92.5\%.

\subsection{When old knowledge stays useful: the mixed stream}\label{sec:mixed}

In the mixed stream the home content returns again and again, so units are worth keeping, and inheritance has something to pass on.
Table~\ref{tab:mixed} gives the verdicts for three inheritance ratios.

\begin{table}[t]
  \centering
  \small
  \setlength{\tabcolsep}{4.5pt}
  \begin{tabular}{@{}llcllll@{}}
  \toprule
   & & Accuracy & \multicolumn{2}{c}{Above the ordinary network of} & Above random & \\
   & $\lambda$ & (\%) & the final size & the peak size & turnover & Verdict \\
  \midrule
  Adam & 0   & 93.21 & $+1.25$ {\scriptsize[$0.92$, $1.57$]} & $+0.82$ {\scriptsize[$0.53$, $1.10$]} & $+4.82$ {\scriptsize[$4.60$, $5.05$]} & passes \\
       & 0.5 & 93.41 & $+1.59$ {\scriptsize[$1.35$, $1.84$]} & $+1.31$ {\scriptsize[$1.19$, $1.43$]} & $+0.15$ {\scriptsize[$0.10$, $0.20$]} & passes \\
       & 1   & 93.16 & $+1.25$ {\scriptsize[$1.18$, $1.33$]} & $+1.12$ {\scriptsize[$0.93$, $1.30$]} & $+0.53$ {\scriptsize[$0.24$, $0.82$]} & passes \\
  \addlinespace[2pt]
  SGD  & 0   & 92.73 & $-1.17$ {\scriptsize[$-1.45$, $-0.89$]} & $-1.17$ {\scriptsize[$-1.34$, $-1.00$]} & $+6.61$ {\scriptsize[$6.44$, $6.79$]} & not shown \\
       & 0.5 & 93.91 & $-0.02$ {\scriptsize[$-0.24$, $0.20$]} & $-0.06$ {\scriptsize[$-0.24$, $0.12$]} & $+0.17$ {\scriptsize[$-0.02$, $0.36$]} & not shown \\
       & 1   & 93.50 & $-0.24$ {\scriptsize[$-0.47$, $-0.00$]} & $-0.42$ {\scriptsize[$-0.65$, $-0.19$]} & $+0.12$ {\scriptsize[$-0.10$, $0.35$]} & not shown \\
  \bottomrule
  \end{tabular}
  \caption{\textbf{Mixed stream} (2,000 tasks; seeds 15--19; paired differences in points with 95\% intervals).
  Accuracy is the mean block score of the last tenth of the home tasks.
  The method passes if it is above both ordinary networks of matched size, each with its own tuned step size.}
  \label{tab:mixed}
\end{table}

\emph{Adam.}
The \GP{} beats both size-matched ordinary networks at all three inheritance ratios, by 1.2 to 1.6 points against the final size and 0.8 to 1.3 points against the peak size, in every seed.
\emph{SGD.}
It does not: it ties the ordinary networks at $\lambda=0.5$ and is slightly below them at $\lambda=0$ and $\lambda=1$.
The highest accuracies of the whole comparison belong to ordinary networks trained with SGD (up to 94.0\%), and the \GP{} under SGD (93.9\%) cannot be told apart from them.
Under SGD the ordinary network barely degrades on this stream, which leaves little for the method to repair; we could not find a schedule of this kind on which it does.

Three further facts belong next to the table.
First, at $\lambda=0.5$ and~1 much of the gain under Adam is also obtained with random turnover; the rule adds 0.15 and 0.53 points to it.
The choice of units matters much less here than at deployment, where it carries the whole gain.
Only at $\lambda=0$, where about 96 units per task are replaced, does a random choice do real damage.
Second, the units that die are the less useful ones here too: after each visit of unrelated content, the least useful third of the units dies 7 to 13 points more often within 200 tasks than the most useful third (under SGD in five of six cases), and with random turnover there is no difference; but half to three quarters of the most useful third die as well.
Third, the method is more disturbed by a visit: while the second sensor visits, its accuracy on the home content drops by 1.9 to 2.4 points under Adam, against 0.4 to 0.7 points for the ordinary network, and it needs 62 to 75 tasks to recover (under SGD up to 472).

\subsection{Conservation is what controls the size}\label{sec:noconservation}

The no-conservation control (Section~\ref{sec:controls}) feeds each unit according to its own use, without the division that makes a layer's intakes sum to the budget.
On the mixed stream under Adam, all 15 confirmation runs grew until a layer filled the storage of 768 units and were stopped, between task 44 and task 930; each of the 24 settings tried on the tuning seeds did the same on at least one seed.
With conservation the same settings run through all 2,000 tasks at 175 to 256 units per layer.
On the pixel-permutation stream the outcome depends on the inheritance ratio: the network shrinks to 49, 59 and 140 units at $\lambda=0$ and grows to the storage limit at $\lambda=0.5$ and~1.
Because the runs without conservation do not complete, we cannot say whether conservation also improves the accuracy.
What the control shows is narrower and firm: conservation is the part of the rule that bounds the size.

\subsection{Two parents remove duplicates}\label{sec:continual-twoparents}

Division at $\lambda>0$ creates a copy, and copies can persist.
Under SGD on the mixed stream, 36\% and 42\% of the established units of layers 2 and 3 at $\lambda=1$ respond almost identically to another unit (correlation of at least 0.9); under Adam at most 9\% do.
With two parents (Section~\ref{sec:twoparents}) a child is not a copy of either, and the share of such duplicates falls to at most 9\% in every setting, and to at most 3\% when each weight is copied from one of the two parents.
The accuracy is higher than with one parent in three of eight settings (by 0.2 to 0.6 points), not distinguishable in five and lower in none.
The networks are also 7\% to 27\% larger, because each parent gives less and can divide again sooner, so we do not attribute the difference to the mechanism alone.
Against ordinary networks of matched size the verdicts are those of Table~\ref{tab:mixed}: passes under Adam in all four settings we ran, by 1.2 to 1.7 points, and none under SGD.

\section{What the nutrient tracks, and where it fails}\label{sec:analysis}

\paragraph{The nutrient level follows the value of a unit.}
The rule never measures what a unit is worth to the loss, so we measured it ourselves.
On the pixel-permutation stream we silenced units one at a time and recorded the rise of the loss.
The rank correlation between this rise and the nutrient level is 0.43 to 0.54 in the three layers at $\lambda=0$ and 0.30 to 0.42 at $\lambda=0.5$.
Removing the best-fed quarter of a layer costs 21.5, 7.9 and 2.7 points of accuracy in layers 1, 2 and 3; removing a random quarter costs 5.3, 2.8 and 1.1 points; removing the worst-fed quarter costs at most 0.65.
A quantity that is routed from the outputs through activities and absolute weights, and that never sees the loss, is thus a usable estimate of which units carry the function.

\paragraph{Useful units die too.}
The estimate is not sharp.
At deployment about 70\% of the pretrained units that mattered for the common content are replaced by the end of the stream (Section~\ref{sec:deploy-which}), and on the mixed stream between half and three quarters of the most useful third die within 200 tasks.
A likely reason is that the death threshold acts on a unit's share within its layer, and after the switch-on with a reduced budget every share is small.
The network keeps its accuracy through this, presumably because the children that replace a unit inherit from well-fed parents; a rule that spared more of the useful units might do better, and we have not looked for one.

\paragraph{Woken units do not hold their place.}
A dormant unit wakes when the intake it would receive reaches the mean of its layer.
Once awake it receives about a tenth of that (median would-be intake 0.47 to 0.63, median real intake after 500 updates 0.04 to 0.08), and 36\% to 58\% of the woken units are below the death threshold again when their protected time ends.
A likely reason is that the test uses the unit's activation on the current batch, while the unit returns with the activity average it had when it fell asleep, which is what the routing reads.
Dormancy keeps rare knowledge in spite of this (Section~\ref{sec:deploy-dormancy}).
A wake rule that lets a returning unit claim its share is the most direct improvement we see.

\paragraph{Why Adam and SGD differ.}
We have no tested explanation, only observations.
Ordinary networks trained with Adam on our streams accumulate inactive units (68\% and 51\% of layers 2 and 3 on the mixed stream, against 32\% and 23\% with SGD; over 90\% at deployment), their output weights grow by a factor of ten, and they adapt more slowly to each new task as the stream goes on; with SGD the adaptation becomes faster.
With the \GP{} the share of permanently inactive units is at most 4\% under both optimisers on the mixed stream.
One reading is that the \GP{} mostly repairs a waste that Adam creates and SGD does not, so that it has more to gain under Adam.
A second observation fits this: under SGD copies of a parent persist (Section~\ref{sec:continual-twoparents}), which suggests that new units are put to different use under the two optimisers.
Both are hypotheses.
We state them because the difference between the optimisers is the main open question of this work.

\paragraph{Cost in time.}
All costs in this paper are counted parameters.
In wall-clock time, our implementation of a network with the \GP{} takes about 1.2 times as long per update as an ordinary network with the same number of live units, after a change to the optimiser bookkeeping that leaves the results bit-identical (measured on one seed and five tasks); the \GP{} itself, with its observations and structural checks, accounts for about a tenth of the time.
The experiments reported here ran at 1.6 to 2.1 times the time of the ordinary network of the starting width.

\section{Related work}\label{sec:related}

\paragraph{Where the routing rule comes from.}
The rule by which a unit passes its nutrient down, \eqref{eq:intake}, is not new.
It has the form of the conservative redistribution of layer-wise relevance propagation~\citep{bach2015pixel} and of excitation backprop~\citep{zhang2016top}, which explain a prediction by passing a class score from the output back towards the input in such a way that every layer holds the same total.
\citet{yeom2021pruning} use such relevances to prune pretrained networks that are adapted to a specialised task.
Other scores that are propagated from the outputs~\citep{yu2017nisp} or conserved across layers~\citep{tanaka2020pruning} have likewise served pruning, with the amount to remove fixed in advance.
In these uses the allocation is a separate step, applied to a network before or after it is trained: an explanation of one prediction, or a pruning pass computed from labelled reference samples and followed by fine-tuning.
The closest exception is the pruning method of Shibata et al.~\citep{shibata2013neural,shibata2015structure}: during backpropagation training, a reward proportional to the current number of neurons is passed from the outputs down to the inputs, split in proportion to $|\text{weight}\times\text{output}|$, and neurons whose leaky accumulated reward falls below a threshold are deleted.
It only removes units, and its reward shrinks with the network; it is a pruning criterion, not a resource that units live on.
We turn the allocation into a growth rule that runs throughout training.
We apply it to running averages of activity and to absolute weights, start it from an equal supply at every output instead of the score of a labelled class, and repeat it every 100 updates as a fixed budget, the same in every layer, that units live on and compete for, with thresholds for division and death.
This change of role, from an analysis or pruning criterion to a growth rule inside training, is what we regard as the main original step of this work.

\paragraph{Growing, pruning and replacing units.}
Networks have been grown by adding units~\citep{fahlman1989cascade,yoon2017lifelong,draelos2017neurogenesis} and by duplicating or splitting them~\citep{chen2015net2net,wei2016network,liu2019splitting,wu2021firefly,evci2022gradmax}; our division at $\lambda=1$ is the widening step of Net2Net applied to one unit.
They have been pruned by magnitude or importance~\citep{han2015learning,hu2016network,frankle2019lottery}, rewired at a fixed number of parameters during training~\citep{mocanu2018scalable,bellec2018deep,evci2020rigging}, grown and pruned in turn~\citep{dai2017nest,dai2018metabolize}, and pruned while a pretrained model is fine-tuned~\citep{sanh2020movement}.
In these methods growth and removal are decided by the loss, by gradients or by magnitudes, on a schedule and towards a size or sparsity that is set in advance.
An exception in single-hidden-layer networks is the sensitivity-based design of Qiao, Han and colleagues~\citep{qiao2010optimal,han2010design}: each hidden unit's share of the output's sensitivity, normalised over the layer, splits the unit when it is high and deletes it when it is low, so that the width is not fixed in advance.
The share is a score recomputed at each check from the single output; it is not routed through several layers, not kept by the units, and not a budget they compete for.

Work on the loss of plasticity replaces or resets units in a network of fixed size.
Continual Backprop~\citep{dohare2021continual,dohare2024loss} reinitialises, at a fixed rate, the mature units of lowest utility, a loss-free measure that is built, like our use, from activations and outgoing weights; its maturity threshold corresponds to our protection period and its reinitialised unit to our child at $\lambda=0$.
Other methods recycle inactive units~\citep{sokar2023dormant}, or perturb, regularise or reset weights~\citep{ash2020warm,nikishin2022primacy,kumar2023maintaining,elsayed2024addressing}.
(\citeauthor{sokar2023dormant} call inactive units dormant; our dormant units are stored units that are out of the computation.)
Against these methods the \GP{} has no replacement rate and no fixed size, children inherit from a parent, and one conserved quantity decides birth and death alike.
\citet{lillo2026stability} find that newly grown units are often ``forward-active yet backward-starved''; our protection period with its larger step size answers the same difficulty, and our woken units show that it is not solved.

\paragraph{Economies of units.}
That units should live on a conserved currency is an old idea in reinforcement learning.
In Holland's bucket brigade~\citep{holland1985properties}, Schmidhuber's neural bucket brigade~\citep{schmidhuber1989local} and Baum's Hayek machine~\citep{baum1999toward}, as described by \citet{kwee2001market}, agents pass money derived from reward along a chain; bankrupt agents are removed, and in Hayek wealthy agents pay to create children.
There the currency is external reward and the agents are rules or classifiers.
Here it is a fixed budget that never reads a reward or a loss, routed through the weights of a network that is trained by backpropagation.
\citet{peng2015developing} let neurons compete for input activation and die without an error signal, in a network that is not trained by gradients.
Neuroevolution has selected single neurons across a population of networks~\citep{moriarty1996efficient}; our selection acts inside one network during its lifetime.

\paragraph{Biology.}
The \GP{} is inspired by the neurotrophic hypothesis and is not a model of it: developing neurons compete for survival factors that their targets supply in limited amounts~\citep{levimontalcini1987nerve,purves1988body,oppenheim1991cell,vanooyen1999competition}, and cells in general die unless other cells signal them to survive~\citep{raff1992social}.

\paragraph{What is new.}
To our knowledge, this is the first learning rule in which one conserved quantity, routed from the outputs with the same budget in every layer, decides when the hidden units of a gradient-trained network divide, survive or die, without reading the loss, the labels or a gradient.
Methods that keep old knowledge by freezing or gating parameters per task~\citep{mallya2017packnet,serra2018overcoming,rusu2016progressive} rely on knowing the task; we are not aware of a counterpart to dormancy with a wake test based on the same nutrient.
Our only baseline is the ordinary network: we have not run any of the methods above and make no claim about how the \GP{} compares with them.

\section{Limitations}\label{sec:limits}

\paragraph{The results depend on the optimiser.}
With Adam the \GP{} is ahead of the ordinary network on the primary metric of all three streams.
With SGD it does not improve accuracy on the mixed stream, and at deployment it trades compute against rarely used content; an ordinary network of the reduced size trained from scratch then keeps the rare content better (Section~\ref{sec:deploy-sgd}).
The strong results under Adam come with a condition: the ordinary Adam network on these streams wastes most of its units and degrades by itself.
Whether large models trained with adaptive optimisers waste units in the same way is an empirical question that we have not addressed.

\paragraph{Scale.}
All networks are perceptrons with three hidden layers, and all data are $28\times 28$ grey-scale images.
The rule is defined for units of a fully connected layer; convolutional channels and attention heads need their own definition of use.
Every comparison rests on five seeds, and we have not repeated the experiments on fresh seeds (Section~\ref{sec:seeds}).

\paragraph{Tuning.}
The structural settings were selected on the mixed stream and reused at deployment without change.
This is a strength of the deployment result, since nothing was fitted to it, and a limit: other thresholds might shift the curves of Figure~\ref{fig:cost} in either direction.
The budget must be chosen below the width, and we know of no rule for the smallest budget that keeps a given accuracy other than trying it.
The \GP{} needs no attention while a network trains, but it is not free of choices made beforehand: the budget, the structural constants and the storage capacity are set by hand, and a run that outgrows its storage has to be discarded.
It also changes the optimiser in one place, the larger step size of newborn units.

\paragraph{What is not conserved, not kept and not measured.}
Conservation holds for the nutrient routed at each check, not for the levels, which dip after many deaths.
Useful units die in large numbers, and woken units rarely hold their place (Section~\ref{sec:analysis}).
Dormancy reduces compute and not storage, and only pretrained units can sleep; letting later units sleep requires a rule for deleting those that never return.
Costs are counted parameters, not measured time or energy.
On the mixed stream the runs without conservation do not complete, so the effect of conservation on accuracy is unknown.

\section*{Acknowledgment}
We disclose that experiment code and analysis scripts were written with the help of an AI coding agent (OpenAI Codex), and that the manuscript text, figures and tables were drafted with the help of an AI assistant (Anthropic Claude). The authors designed the study, directed the experiments, verified all reported numbers against the experiment records, and take full responsibility for the content.

\bibliographystyle{abbrvnat}
\bibliography{refs,refs_related}

\appendix
\section{Method details}\label{app:method}

\paragraph{Protection and capacity.}
A newborn unit starts with age zero.
It can neither die nor divide for $P$ updates, and during this period the step size of its incoming weights and bias is multiplied by $\kappa\ge 1$.
Under Adam, the moment estimates of its connections start at zero, with a bias correction counted from its birth.
In a network trained from scratch all initial units start with age zero and level $B/N_\ell$, so they too are protected, and use the multiplied step size, for the first $P$ updates.
Storage holds at most $C$ units per layer; a run in which a layer reaches $C$, loses all its units or produces non-finite values is invalid and is not used for a selection or a comparison.

\paragraph{Settings.}
The constants above form a \emph{structural setting}; settings are selected on tuning seeds (Section~\ref{sec:seeds}) and given with the results.
All deployment experiments use $\thdie=0.05$, $\thdiv=1.5$, $\tau=P=3{,}750$ updates, $\kappa=3$, $\varepsilon=0.1$, $\rho=0.01$, $\lambda=0.5$ and $C=768$.

\subsection{Division with two parents}\label{app:twoparents}

In the two-parent variant the first parent $p$ is chosen as before, and a second parent $q$ is drawn among the other established units of the layer with probability proportional to their levels.
In \eqref{eq:child}, $w_p$ is replaced by a combination of $w_p$ and $w_q$: either each entry (each incoming weight, and the bias) is copied from one of the two parents with equal probability, or the two vectors are averaged.
Each parent gives a quarter where a single parent gives a half: the child receives $\tfrac{\lambda}{4}(W_{pj}+W_{qj})$, each parent keeps $1-\tfrac{\lambda}{4}$ of its outgoing weights, and the levels move in the same proportions, so the total level of the layer is again unchanged.
Everything else stays as it is; if a layer has no second established unit, the division falls back to one parent.
The child is in general not a copy of either parent, so for $\lambda>0$ the function of the network changes at the division.

\subsection{Dormancy instead of death}\label{app:dormancy}

Death is irreversible: a unit that is needed only rarely can starve between two uses, and what it has learned is then lost.
For a pretrained network we therefore also study a variant in which starving pretrained units are put to sleep.
We call the units that are present when the \GP{} is switched on the \emph{founder units}.
An established founder unit whose level falls below $\thdie$ becomes \emph{dormant} where it would otherwise die.
It takes no part in the forward or the backward pass; its weights, bias, optimiser state and \GP{} state are kept unchanged in a separate store, and its slot stays reserved.
Units born after the switch-on die as before.

At every check, each dormant unit $d$ of layer $\ell$ is asked what it would receive if it alone were added back.
Its would-be activity $\hat a_d$ is its mean activation on the last training batch, computed from its stored incoming weights and the current activations of the layer below.
Its would-be intake follows from \eqref{eq:intake} with $d$ added to the live units,
\begin{equation}
  \hat n_d = \sum_{j} \sigma_j\,\frac{\hat a_d\,|W_{dj}|}{\sum_{k\in\live_\ell} u_{kj} + \hat a_d\,|W_{dj}|} ,
  \label{eq:wake-app}
\end{equation}
with its stored outgoing weights $W_{dj}$ and the supplies $\sigma_j$ of this check.
The unit wakes if $\hat n_d \ge B/N_\ell$, the mean intake of the live units of its layer; the test adds no parameter.
A woken unit returns with its stored weights and state and competes like any other unit, except that it cannot go dormant again for one protection period (3,750 updates).
Because its stored level returns with it, the total level of a layer can briefly exceed $B$ after wakes; the nutrient routed at each check stays conserved.
Only founder units can go dormant in this paper: extending dormancy to units born later needs a rule for deleting units that have slept for long, which we leave open.

\subsection{Counting cost}\label{app:cost}

We count cost in parameters that take part in the computation.
The \emph{compute cost} of a network is the number of weights and biases between its live units, divided by 99,710, the parameter count of the 784--100--100--100--10 network; the ordinary 784--300--300--300--10 network costs 4.2033.
With dormant units the compute cost includes the wake test: for each dormant unit its connections from and to live units and its bias, counted once per check, that is, divided by $K$.
The \emph{storage cost} counts live and dormant units together; without dormancy it equals the compute cost.
The count leaves out the bookkeeping of the \GP{} itself and is not a wall-clock measurement (Section~\ref{sec:limits}).

\section{Protocol details}\label{app:protocol}

The comparisons, metrics and decision rules below were fixed before the method was run on the seeds used for conclusions.
Sections~\ref{sec:deployment} and~\ref{sec:continual} report against them.

\subsection{Primary baseline: an ordinary network}\label{app:baseline}

The baseline throughout is an \emph{ordinary network}: the same multilayer perceptron with a fixed structure, trained by backpropagation on the same stream with the same optimiser.
Its step size is tuned on tuning seeds, separately for every size at which it is used (on the mixed stream and the deployment stream over 12 logarithmically spaced points, from $10^{-4}$ to $10^{-2}$ for Adam and from $10^{-2}$ to $1$ for SGD).
A result counts in favour of the method only relative to this network.

When a network is trained from scratch, the method chooses its own size, so the ordinary network is run at two matched sizes: the \emph{final size} of the method (its mean number of live units per layer over the evaluation window) and its \emph{peak size} (the largest size it reaches in the stream).
The method has to beat both.
In the deployment stream the baseline is the same pretrained network, trained on in the ordinary way at its fixed size.

\subsection{Three controls}\label{app:controls}

\emph{Random turnover} replays the real run of the same seed: at every check and in every layer it applies the same number of deaths and divisions (with dormancy, also of sleeps and wakes), but draws the units uniformly at random among the units eligible for the event.
It separates \emph{which} units are replaced from \emph{how many} are replaced and \emph{when}.

\emph{No conservation} changes one thing: unit $i$ receives $c_\ell\,\sigma_j\,u_{ij}$ from downstream unit $j$, without the division by the total use in \eqref{eq:intake}.
The constant $c_\ell$ is fixed once per layer, at the end of the initial protection period, such that the intakes of the layer sum to $B$ at that moment.
From then on the intake of a unit depends on its own use only, and the total of a layer is free to rise or fall.

A \emph{same-size ordinary network trained from scratch} has the widths that the method ends with and its own tuned step size, and is trained on the whole stream.
It asks whether a result is explained by the size alone.
On the two streams trained from scratch it is the baseline itself.
On the deployment stream it is a reference only: it needs the target size in advance and training from scratch, and neither is available when a pretrained model is deployed.

\subsection{Tuning seeds, check seeds and rules fixed in advance}\label{app:seeds}

Seeds 0--4 are \emph{tuning seeds}.
Every choice is made on them by a fixed rule, for the method and for each ordinary network alike: among the valid settings (no invalid run on any tuning seed, Section~\ref{sec:growthpath}), the one with the highest mean primary metric is selected.
The settings of the method count towards its tuning budget.
The selection is frozen before any check seed is trained.
Each comparison is then run once on five \emph{check seeds} that no tuning run has used: seeds 5--9 and 10--14 on the pixel-permutation stream, and seeds 15--19 on the mixed stream and the deployment stream.
On the deployment stream nothing of the method was tuned: the structural setting and the inheritance ratio are those selected on the mixed stream, and the step size is that of the pretrained network, so that the \GP{} is the only difference from the baseline.

Stream lengths, primary metrics and verdict thresholds were written down before the method was run on the check seeds; readings that were not fixed in advance are marked as such where they appear.
We have not run a further confirmation on fresh seeds: every result in this paper rests on the five check seeds of its comparison.

For a given seed, all networks see the same stream.
On the streams trained from scratch the method starts from the initial weights of the ordinary network of its starting width; at deployment every method starts from the same pretrained network, including its optimiser state.
We use two optimisers, SGD without momentum and Adam at its default settings, with batch size 16, no weight decay and no schedule.
Comparisons are always between runs with the same optimiser, and the tables show the two side by side with a separate verdict for each; where a comparison was run with one optimiser only, we say so.
All comparisons are paired by seed: we report the mean of the five per-seed differences with its 95\% interval from Student's $t$ distribution with 4 degrees of freedom, and we call a difference positive or negative only if this interval excludes zero.

\subsection{Three task streams}\label{app:streams}

A stream is a sequence of ten-class classification tasks that follow one another without any signal at the boundary; every training image of a task is seen once.
All networks have 784 inputs, three hidden layers and 10 outputs, and the time constants of the \GP{} are kept in updates on all streams.

\paragraph{Pixel-permutation stream.}
Each task is the 60,000 training images of MNIST~\citep{lecun1998gradient} under a new random permutation of the pixels~\citep{goodfellow2013empirical}, in a new random order (3,750 updates).
The stream has 200 tasks, and 800 in a long version; networks start with 100 units per layer (300 in a second series), and $B$ is the starting width.
The score is the online accuracy: every batch is predicted before it is used for training.
The primary metric is the mean online accuracy over the last tenth of the tasks (tasks 181--200).
Every task rewires the input, so this stream tests whether a network can keep learning; it gives no credit for keeping what was learned.

\paragraph{Mixed stream.}
The stream has 2,000 tasks.
Each task has 24,000 training images (1,500 updates) of ten classes, which are assigned to the ten output units anew in every task.
The \emph{home} kinds are handwritten digits (MNIST), handwritten letters (ten of the 37 letter classes of EMNIST Balanced~\citep{cohen2017emnist}) and Kuzushiji characters (KMNIST~\citep{clanuwat2018kmnist}); two successive home tasks are never of the same kind.
Two unrelated kinds visit once and never return: clothing (Fashion-MNIST~\citep{xiao2017fashion}) in tasks 401--600, and a \emph{second sensor} (home images under one fixed pixel permutation) in tasks 1,201--1,400.
During a visit each task is of the visiting kind with probability one half.
The score of a task is its \emph{block score}: the mean accuracy on 4,000 held-out test images (400 per class), measured after 6,000, 12,000, 18,000 and 24,000 training images of the task.
The primary metric is the mean block score of the last tenth of the home tasks.
Networks start with 100 units per layer, and $B=100$.

The mixed stream was selected among six candidate schedules with ordinary networks only, by rules fixed in advance, for example that the ordinary network declines by at least one point along the stream.
The selection was decided under Adam; under SGD no candidate met that rule, and Section~\ref{sec:continual} returns to this.

\paragraph{Deployment stream.}
Tasks have the format of the mixed stream.
In the \emph{general phase} (tasks 1--500) each task is drawn with equal probability from five kinds: digits, letters, Kuzushiji, clothing and the second sensor.
In the \emph{deployment phase} (tasks 501--2,000) the content falls into three groups.
Digits and letters are the \emph{common} content.
Kuzushiji is the \emph{rare} content: exactly one task in every block of 20, at a random position.
Clothing and the second sensor are \emph{gone}: they never appear again.
A network with three hidden layers of 300 units is trained through the general phase as an ordinary network.
At the end of task 500 the \GP{} is switched on in this network, with a budget $B$ between 25 and 100 (and with $B=300$, the case without a reduced budget) and with every unit established (Section~\ref{sec:budget}).
The baseline is the same network trained on without it.

\subsection{Metrics and verdict rules}\label{app:verdicts}

On the streams trained from scratch, the method passes if its primary metric is higher than that of both same-size ordinary networks, with both 95\% intervals above zero.
Next to every accuracy we report the compute cost of Section~\ref{sec:cost}, averaged over all updates of the stream (\emph{training cost}) and over the evaluation window (\emph{deployed cost}).

For deployment we use four numbers.
The \emph{common-content accuracy} is the mean block score of the last tenth of the common tasks of the deployment phase (142 tasks).
The \emph{late rare-content accuracy} is the mean block score of the 25 Kuzushiji tasks among tasks 1,501--2,000.
The \emph{compute cost} and the \emph{storage cost} are averaged over the updates of the common-content window.
The verdict rules are:
\begin{enumerate}[nosep,leftmargin=2.2em,label=(\alph*)]
  \item \emph{not worse on the common content}: the lower end of the interval of the difference to the ordinary network is above $-0.5$ points;
  \item \emph{smaller}: the upper end of the interval of the compute cost is at most 75\% of the cost of the ordinary network, that is, at most 3.1525 in units of the 100-wide network;
  \item \emph{rare content kept}: the lower end of the interval of the late rare-content difference to the ordinary network is above $-1.0$ point.
\end{enumerate}
Rules (a) and (b) were fixed before the method was run on this stream; the rule with death is judged on them, with its rare-content accuracy reported next to them.
Rule (c) was added after the first results with death and before dormancy was run; dormancy is judged on all three.

\section{Additional tables}\label{app:tables}
\begin{table}[h]
  \centering
  \small
  \begin{tabular}{@{}lrllccc@{}}
\toprule
 & Compute & Common content & Late rare content & \multicolumn{3}{c}{Rules} \\
Network & (\%) & (points) & (points) & (a) & (b) & (c) \\
\midrule
\multicolumn{7}{@{}l}{\textbf{Adam}} \\
Ordinary network & 100 & 93.92 & 72.41 & & & \\
\addlinespace[2pt]
Growth Path, $B=100$ & 60 & $+1.16$ {\scriptsize[$0.76$, $1.55$]} & $+4.3$ {\scriptsize[$1.1$, $7.5$]} & \checkmark & \checkmark & \checkmark \\
Growth Path, $B=80$ & 43 & $+1.14$ {\scriptsize[$0.84$, $1.44$]} & $+3.6$ {\scriptsize[$0.2$, $7.0$]} & \checkmark & \checkmark & \checkmark \\
Growth Path, $B=65$ & 33 & $+1.13$ {\scriptsize[$0.78$, $1.48$]} & $+2.7$ {\scriptsize[$-0.2$, $5.7$]} & \checkmark & \checkmark & \checkmark \\
Growth Path, $B=50$ & 24 & $+1.01$ {\scriptsize[$0.69$, $1.33$]} & $+0.8$ {\scriptsize[$-2.7$, $4.3$]} & \checkmark & \checkmark & $\times$ \\
Growth Path, $B=35$ & 17 & $+0.81$ {\scriptsize[$0.44$, $1.18$]} & $-0.9$ {\scriptsize[$-4.5$, $2.6$]} & \checkmark & \checkmark & $\times$ \\
Growth Path, $B=25$ & 12 & $+0.54$ {\scriptsize[$0.21$, $0.86$]} & $-3.5$ {\scriptsize[$-7.0$, $-0.1$]} & \checkmark & \checkmark & $\times$ \\
\addlinespace[2pt]
\quad with dormancy, $B=100$ & 71 & $+1.20$ {\scriptsize[$0.87$, $1.53$]} & $+8.1$ {\scriptsize[$5.1$, $11.2$]} & \checkmark & \checkmark & \checkmark \\
\quad with dormancy, $B=80$ & 53 & $+1.18$ {\scriptsize[$0.85$, $1.51$]} & $+7.6$ {\scriptsize[$3.9$, $11.2$]} & \checkmark & \checkmark & \checkmark \\
\quad with dormancy, $B=65$ & 41 & $+1.11$ {\scriptsize[$0.71$, $1.50$]} & $+6.7$ {\scriptsize[$3.8$, $9.6$]} & \checkmark & \checkmark & \checkmark \\
\quad with dormancy, $B=50$ & 31 & $+1.04$ {\scriptsize[$0.63$, $1.44$]} & $+5.4$ {\scriptsize[$2.2$, $8.5$]} & \checkmark & \checkmark & \checkmark \\
\quad with dormancy, $B=35$ & 22 & $+0.84$ {\scriptsize[$0.55$, $1.13$]} & $+3.8$ {\scriptsize[$1.3$, $6.4$]} & \checkmark & \checkmark & \checkmark \\
\quad with dormancy, $B=25$ & 16 & $+0.62$ {\scriptsize[$0.28$, $0.97$]} & $+1.1$ {\scriptsize[$-1.5$, $3.7$]} & \checkmark & \checkmark & $\times$ \\
\midrule
\multicolumn{7}{@{}l}{\textbf{SGD}} \\
Ordinary network & 100 & 96.15 & 87.84 & & & \\
\addlinespace[2pt]
Growth Path, $B=100$ & 77 & $+0.31$ {\scriptsize[$0.16$, $0.45$]} & $-1.5$ {\scriptsize[$-2.0$, $-1.0$]} & \checkmark & $\times$ & $\times$ \\
Growth Path, $B=80$ & 57 & $+0.20$ {\scriptsize[$0.11$, $0.29$]} & $-2.4$ {\scriptsize[$-2.9$, $-1.9$]} & \checkmark & \checkmark & $\times$ \\
Growth Path, $B=65$ & 42 & $+0.08$ {\scriptsize[$-0.06$, $0.21$]} & $-3.5$ {\scriptsize[$-3.9$, $-3.0$]} & \checkmark & \checkmark & $\times$ \\
Growth Path, $B=50$ & 31 & $-0.05$ {\scriptsize[$-0.17$, $0.07$]} & $-5.0$ {\scriptsize[$-5.5$, $-4.4$]} & \checkmark & \checkmark & $\times$ \\
Growth Path, $B=35$ & 21 & $-0.40$ {\scriptsize[$-0.51$, $-0.28$]} & $-7.4$ {\scriptsize[$-8.2$, $-6.5$]} & $\times$ & \checkmark & $\times$ \\
Growth Path, $B=25$ & 15 & $-0.85$ {\scriptsize[$-0.92$, $-0.78$]} & $-11.7$ {\scriptsize[$-12.8$, $-10.5$]} & $\times$ & \checkmark & $\times$ \\
\addlinespace[2pt]
\quad with dormancy, $B=100$ & 85 & $+0.31$ {\scriptsize[$0.20$, $0.42$]} & $-0.3$ {\scriptsize[$-0.7$, $0.1$]} & \checkmark & $\times$ & \checkmark \\
\quad with dormancy, $B=80$ & 64 & $+0.22$ {\scriptsize[$0.13$, $0.30$]} & $-0.9$ {\scriptsize[$-1.6$, $-0.2$]} & \checkmark & \checkmark & $\times$ \\
\quad with dormancy, $B=65$ & 51 & $+0.11$ {\scriptsize[$0.03$, $0.18$]} & $-1.6$ {\scriptsize[$-2.2$, $-1.0$]} & \checkmark & \checkmark & $\times$ \\
\quad with dormancy, $B=50$ & 39 & $+0.04$ {\scriptsize[$-0.09$, $0.17$]} & $-2.8$ {\scriptsize[$-3.4$, $-2.2$]} & \checkmark & \checkmark & $\times$ \\
\quad with dormancy, $B=35$ & 27 & $-0.17$ {\scriptsize[$-0.31$, $-0.04$]} & $-5.5$ {\scriptsize[$-5.8$, $-5.1$]} & \checkmark & \checkmark & $\times$ \\
\quad with dormancy, $B=25$ & 19 & $-0.62$ {\scriptsize[$-0.71$, $-0.53$]} & $-8.5$ {\scriptsize[$-9.1$, $-8.0$]} & $\times$ & \checkmark & $\times$ \\
\bottomrule
\end{tabular}

  \caption{Deployment, all six budgets (seeds 15--19; columns and rules as in Table~\ref{tab:deploy}).}
  \label{tab:deploy-full}
\end{table}

\end{document}